%% file: jmlr.tex
 \documentclass[pmlr,twocolumn,final,10pt]{jmlr} % W&CP article

\makeatletter
\renewcommand*\@jmlrpages{} % Redefine \@jmlrpages to be empty
\makeatother

\usepackage{booktabs}
\usepackage{siunitx}

\usepackage[switch]{lineno}

\usepackage[acronym]{glossaries}
\makeglossaries
\glsdisablehyper

\usepackage{dsfont}
\theorembodyfont{\upshape}
\theoremheaderfont{\scshape}
\theorempostheader{:}
\theoremsep{\newline}

\jmlrworkshop{Machine Learning for Health (ML4H) 2024} % W&CP title

\title[Integrated Machine Learning \& Survival Analysis Modeling for Enhanced CKD Risk Stratification]{Integrated Machine Learning and Survival Analysis Modeling for Enhanced Chronic Kidney Disease Risk Stratification}

\author{%
  \Name{Zachary Dana} \Email{zdana@vironix.ai}\\
  \Name{Ahmed Ammar Naseer} \Email{aanaseer22@gmail.com}\\
  \Name{Botros Toro} \Email{btoro@vironix.ai}\\
  \Name{Sumanth Swaminathan} \Email{sswami@vironix.ai}\\
  \addr Vironix Health Inc, Austin, TX, USA
}

\begin{document}
\include{glossary}

\maketitle

\input{0-abstract-data-avail-IRB}

\input{1-introduction}

\input{4-method}

\input{5-experiments}

\input{6-results-discussion}

\input{7-limitations}

\input{8-conclusion}

% \acks{Acknowledgments go here \emph{but should only appear in the camera-ready version of the paper if it is accepted}. Acknowledgments do not count toward the paper page limit.}

\bibliography{jmlr}

\appendix

\input{9-appendix-A-figure}
\input{9-appendix-B-related-work}
\input{9-appendix-B1-kfre}
\input{9-appendix-C-binary-classifier}
\input{9-appendix-C2-shap}
\input{9-appendix-C3-cphm}
\input{9-appendix-D-eval-metrics}
\input{9-appendix-E-supp-tables}
\input{9-appendix-F-mean-shap-val}
\input{9-appendix-G-brier}
\input{9-appendix-H-dynamic-auc}

\end{document}

%% file: glossary.tex
% Numeric

% A
\newacronym{auroc}{AUROC}{area under receiving operating characteristic curve}

% B

% C
\newacronym{ckd}{CKD}{chronic kidney disease}
\newacronym{cphm}{CPHM}{cox proportional hazards model}
\newacronym{c-index}{C-index}{concordance Index}

% D
\newacronym{dt}{DT}{decision tree}
\newacronym{dca}{DCA}{decision curve analysis}

% E
\newacronym{egfr}{eGFR}{estimated glomerular filtration rate}
\newacronym{ehr}{EHR}{electronic health record}
\newacronym{esrd}{ESRD}{end-stage renal disease}

% F
\newacronym{fcnn}{FCNN}{fully connected neural network}

% G

% H

% I

% J

% K
\newacronym{kfre}{KFRE}{kidney failure risk equation}

% L
\newacronym{lr}{LR}{logistic regression}

% M
\newacronym{ml}{ML}{machine learning}
\newacronym{mchc}{MCHC}{mean corpuscular hemoglobin concentration}
\newacronym{mch}{MCH}{mean corpuscular hemoglobin}

% N

% O

% P

% Q

% R
\newacronym{relu}{ReLU}{rectified linear unit}
\newacronym{resnet}{ResNet}{residual neural network}
\newacronym{rf}{RF}{random forest}

% S
\newacronym{shap}{SHAP}{shapley additive explanations}

% T

% U
\newacronym{uacr}{UACR}{urine albumin-to-creatinine ratio}

% V

% W

% X
\newacronym{xgboost}{XGBoost}{extreme gradient boosting}

% Y

% Z

%% file: 0-abstract-data-avail-IRB.tex
\begin{abstract}
Chronic kidney disease (CKD) is a significant public health challenge, often progressing to end-stage renal disease (ESRD) if not detected and managed early. Early intervention, warranted by silent disease progression, can significantly reduce associated morbidity, mortality, and financial burden. In this study, we propose a novel approach to modeling CKD progression using a combination of machine learning techniques and classical statistical models. Building on the work of \citet{liu2023combining}, we evaluate linear models, tree-based methods, and deep learning models to extract novel predictors for CKD progression, with feature importance assessed using Shapley values. These newly identified predictors, integrated with established clinical features from the Kidney Failure Risk Equation, are then applied within the framework of Cox proportional hazards models to predict CKD progression.

% Our contributions are threefold: (1) Identification of novel predictors for CKD progression through the integration of machine learning and survival models; (2) Demonstration of improved predictive accuracy for CKD progression, as evidenced by higher Concordance Index and lower Brier scores; and (3) Extension of \citet{liu2023combining}  methodology.

% This study presents a novel approach to modeling CKD degeneration risk, integrating machine learning models with classical statistical techniques. Leveraging machine learning models trained on $ \approx 1.4 \text{k} $ input features from the MIMIC-IV clinical database, we employed Shapley Additive exPlanations (SHAP) to identify the top 40 ranked risk factors in each model. We compared “augmented” Cox models incorporating ML-selected novel predictors with a “baseline” Cox model that relied solely on established Kidney Failure Risk Equation (KFRE-8) factors. This approach identified both established and novel predictors, and yielded C-Index improvements of $ \approx 0.0005 $ and $ \approx 0.0011 $ in the XGBoost-augmented and ResNet-augmented Cox models, respectively.
\end{abstract}

\paragraph*{Data and Code Availability}
Code utilized in this study is available as supplemental material. We make use of the publicly available MIMIC-IV (Medical Information Mart for Intensive Care, Version IV) clinical database \citep{mimiciv2024} in our experiments. MIMIC-IV consists of 364,627 thousand medical records for patients admitted in critical care units of the Beth Israel Deaconess Medical Center in Boston, MA from 2008 to 2022. It is an update of the earlier MIMIC-III database \citep{mimiciv2024}. %including data such as vital signs, medications, laboratory measurements, notes from healthcare providers, diagnostic codes, imaging reports, survival data, and other clinically relevant recordings 

% We make use of the publicly available MIMIC-IV clinical database \citet{mimiciv2024}, as well as two supplementary datasets: emergency department (ED) visits data used by \citet{hong2018predicting}, which we refer to as ED EHR, and a clinical database excerpt utilized by \citet{iimori2014prognosis}, which we refer to as CKD-ROUTE.

% \textbf{ED EHR} \quad The ED data consists of deidentified records of 560,486 adult patient visits to three EDs within the Yale New Haven Health system between March 2014 and July 2017 \citet{hong2018predicting}. We select a subset of patient features from these, which include demographics, hospital usage statistics, chief complaint, past medical history, and medications. Data pre- and post-processing is discussed in Appendix C.

% \textbf{CKD-ROUTE} \quad The CKD-ROUTE data consists of 1138 participants from Chronic Kidney Disease Research of Outcomes in Treatment and Epidemiology (CKD-ROUTE) a prospective, observational cohort study of a representative Japanese population with stage G2–G5 CKD, according to the Kidney Disease Improving Global Outcomes (KDIGO) classification, who were not undergoing dialysis \citet{iimori2014prognosis}. All participants visited either the Tokyo Medical and Dental University Hospital or one of its 15 affiliated hospitals in the Tokyo metropolitan area of Japan every 6 months for collection of blood and urine samples to measure clinical variables.

\paragraph*{Institutional Review Board (IRB)}
This paper uses the publicly available and de-identified MIMIC-IV dataset, which does not require IRB approval.

%% file: 1-introduction.tex
\section{Introduction}
\label{sec:intro}
\Gls{ckd} is a major public health concern, characterized by silent disease progression that can lead to \gls{esrd} and the need for kidney transplantation if not detected early \citep{bai2022machine}. \gls{ckd} affects millions globally and is associated with significant morbidity, mortality, and healthcare costs \citep{kerr2012estimating}. The gradual decline in kidney function in \gls{ckd} patients often goes unnoticed until the disease has advanced to a critical stage \citep{kalantar2021chronic}. Early intervention in \gls{ckd} can improve the quality of life and reduce healthcare expenses by slowing disease progression \citep{kalantar2021chronic}.

\newacronym{ml}{ML}{machine learning}

Predictive models and dynamic risk stratification algorithms can enable identification of patients at high risk of \gls{ckd} degeneration. These models facilitate timely modifications in patient management, including adjustments to medication, diet, sleep, and exercise regimens, thereby improving patient outcomes \citep{xiao2019comparison}. In recent years, \gls{ml} techniques have shown great promise in healthcare applications, providing powerful tools for predictive modeling and risk stratification (Appendix \ref{apd:related}).

In this study, we evaluate an approach to modelling the progression of \gls{ckd} stages using a combination of \gls{ml} techniques and classical statistical models. By extending the work proposed by \citet{liu2023combining} we evaluate the use of linear models, tree-based models, and deep learning models in extracting novel predictors for \gls{ckd} degeneration, determined by Shapley values. These novel predictors are then used in conjunction with clinically established \gls{kfre} features (Appendix \ref{apd:kfre}) to model the progression of \gls{ckd} using \glspl{cphm}.

\paragraph{Contributions.}\textbf{1.} Identification of potential novel predictors for CKD progression: By leveraging machine learning models, Shapley value analysis, and classical survival models, this work identifies new features beyond the established clinical predictors in the \gls{kfre}-8 model to aid in predicting CKD progression. \textbf{2.} Improved predictive performance for CKD progression: The integration of machine learning-derived predictors with classical Cox proportional hazards models leads to improved predictive accuracy, as demonstrated by higher C-index and lower Brier scores. \textbf{3.} Extension of \citet{liu2023combining} methodology: by exploring additional models for feature selection and extending the Cox proportional hazards modeling to CKD progression.

%% file: 4-method.tex
\section{Method}
\label{sec:methods}
\subsection{Pipeline}
Our approach follows a structured pipeline for feature selection and modeling. Let $\mathcal{D} = \{\mathbf{x}_i\}_{i=1}^{M}$, denote a dataset where each $\mathbf{x}_{i} \in \mathbb{R}^{N}$ represents a feature vector. Let $f : \mathbb{R}^{N} \mapsto \{0, 1\}$ represent a binary classifier (Appendix \ref{apd:binary_class}). We first train the classifier $f(\cdot;\mathbf{\theta}):\mathbb{R}^{n} \mapsto \{0,1\}$ on $\mathcal{D}^{Tr} \in \mathbb{R}^{P \times n}$ a training data subset from $\mathcal{D}$. After training, we compute the Shapley $\phi_{i}$ for each feature $x_{i}$ (Appendix \ref{apd:shap}). The top $j$ features with the highest mean Shapley values are selected, forming the set $F_{s}$. Next, we combine $F_{s}$ with the \gls{kfre}-8 feature set $F_{\text{KFRE-8}}$ by taking their union, defined as $F = F_{s} \cup F_{\text{KFRE-8}}$. The dataset $\mathcal{D}$ is then reduced to include only features in $F$. Finally, a \gls{cphm} is trained on this reduced data set to predict \gls{ckd} progression (Appendix \ref{apd:cphm}), evaluating the novel predictors' efficacy within the framework of survival analysis. \autoref{fig:ml_feature_selection} illustrates the modelling pipeline used here. This paper expands the methodology proposed by \citet{liu2023combining} through the incorporation of a diverse range of model architectures in the feature selection stage, and extends the application to \gls{ckd}.

%% file: 5-experiments.tex
\section{Experiments}
\label{sec:experiments}
\subsection{Data and data handling}
\paragraph{MIMIC-IV.} We extracted the complete subset of patients from MIMIC-IV with documented \gls{ckd} diagnoses, as indicated by seven relevant ICD-9 codes descibed in \autoref{tab:ckd_icd9}, yielding a cohort of 14,012 patients. Using this subset, we define a binary variable to indicate \gls{ckd} progression, deemed observed if a patient is diagnosed with a more advanced \gls{ckd} stage at any time after an earlier diagnosis of a less severe stage. Among the patients in the cohort, 1,483 (10.6\%) experienced \gls{ckd} stage progression over a median follow-up period of 111.5 days (IQR 6.0-910.25). The feature selection models, were trained on a high-dimensional input set, including a combined 1373 demographic, diagnostic, and lab recording features; the full characteristics of the former are summarized in \autoref{tab:ml_in}.

\subsection{Feature selection}
\subsubsection{Models}
We evaluate the performance of five binary classifiers for the feature selection component of the pipeline: \gls{lr}, \gls{dt}, \gls{rf}, \gls{xgboost}, \glspl{fcnn}, and \gls{resnet}. The linear models (Appendix \ref{apd:linear_models}) and tree-based methods (Appendix \ref{apd:tree_methods}) were trained with the logarithmic loss function as the objective criterion, and hyperparameters were selected through Bayesian optimization; detailed search spaces and optimized values for the latter are provided in \autoref{tab:bayes}. For logistic regression, the solver was specified a \texttt{lbfgs}, and maximum iterations set to 1000.

The neural networks (Appendix \ref{apd:neural_networks}) utilize binary cross-entropy loss with logits and the Adam optimizer. The \gls{fcnn} model was configured with 4 hidden layers containing 512, 256, 128, and 1 neuron, each followed by a dropout layer with the rate set to 0.2. For \gls{resnet}, we used 3 residual blocks, each consisting of 2 fully-connected layers with a hidden dimension of 64. Single fully-connected layers were applied prior to and following the residual blocks, yielding an architecture with 8 layers in total. Rectified linear unit (\acrshort{relu}) activation was applied in both architectures. The learning rate, weight decay (L2 regularization), maximum epoch, and early stopping hyperparameters are reported in \autoref{tab:nn_hparams}.

All models were trained and evaluated across five distinct data splits, utilizing unique random seeds for cross-validation. For the linear and tree-based methods, the training sets were further divided into five validation folds to support Bayesian optimization, prior to applying the tuned hyperparameter values in model cross-validation with the complete data splits.

\subsubsection{Features}
For each binary classifier, we identified the 40 features with the highest mean absolute Shapley values. The union of these top features, along with those defined in \gls{kfre}-8, was then selected as the final feature set. Because of the computational infeasibility of calculating Shapley values as defined in equation \eqref{eqn_shapley_values} for the large number of features, we utilized \gls{shap} \citep{lundberg2017shap} to approximate these values. Specifically, \gls{shap} values were computed using the TreeExplainer \citep{lundberg2020local2global} for tree-based models, the LinearExplainer for linear models, and the DeepExplainer for deep learning models. 

\subsection{Cox proportional hazards model}
Following the feature selection, \glspl{cphm} were fitted to explore the associations between the identified novel predictors and \gls{ckd} stage progression. The data for the final feature set obtained from each binary classifier were used to train the \glspl{cphm}. Additionally, a baseline \gls{cphm} was trained using only the \gls{kfre}-8 features to serve as a control. The \glspl{cphm} were implemented with a penalizer set to 0.0007, the minimum value necessary to prevent overfitting. Model fitting involved five-fold cross-validation, and the optimal model was selected based on the highest average \gls{c-index} (Appendix \ref{subsec:cindex}) across the validation sets. The models with the best performance, as indicated by the highest average \gls{c-index}, were preserved for further analysis, along with the corresponding training and testing datasets. The proportional hazards assumption was evaluated using Schoenfeld residuals. Additionally, Brier score (Appendix \ref{subsec:brier}) and dynamic \gls{auroc} plots (Appendix \ref{subsec:auroc}) were generated for the full set of hazards models, offering an assessment of the alignment between predicted risks and observed outcomes over time.

We computed Brier score at 5-year for assessing overall model performance and \gls{c-index} for assessing risk discrimination. We used the two sets of baseline hazard at 5-year and model coefficients (i.e. “beta values”) yielded from these two Cox models to compute the 5-year risks and prognostic index (i.e. variable beta) of each participant in the training (80\%) and test (20\%) data, respectively. These 5-year risks were subsequently used to compute Brier score at 5 years and prognostic indices were used to compute \gls{c-index} using the training and test data, respectively.

%% file: 6-results-discussion.tex
\section{Results and Discussion}\label{sec:discussion}

\subsection{Feature selection results}

\paragraph{\gls{auroc}.} \autoref{tab:ml_model_auc} shows the \gls{auroc} scores obtained by the \gls{ml} models used in the feature extraction process. The comparison of models based on \gls{auroc} shows that \gls{xgboost} performs the best, with the highest average \gls{auroc} of 0.7796 and a best score of 0.8105. \gls{dt} follows with an average score of 0.7283, and the best-performing model scores 0.7799. \gls{lr} achieves an average \gls{auroc} of 0.7027, performing slightly below \gls{dt}. \Gls{rf} performs the worst, with an average of 0.5796. Deep learning models such as \gls{fcnn} and \gls{resnet} perform similarly, with average \gls{auroc} values of 0.6612 and 0.6540, respectively.

\paragraph{\gls{shap}.} Figures \ref{fig:xgboost_shapma}, \ref{fig:fcnn_shapma}, \ref{fig:resnet_shapma}, \ref{fig:lr_shapma}, \ref{fig:dt_shapma}, and \ref{fig:rf_shapma} show the top 40 mean absolute \gls{shap} values obtained by the feature selection models. Across nearly all models, features related to creatinine (e.g., creatinine mean, max, last, and median) consistently rank among the most important predictors. This trend is particularly observed in the \gls{xgboost}, \gls{rf}, \gls{lr}, and \gls{dt} models. Renal dialysis status appears as one of the top contributors in almost all models, particularly in \gls{resnet}, \gls{lr}, \gls{fcnn}, and \gls{dt}. Urea nitrogen frequently appears in the top set of important features, particularly in the \gls{xgboost}, \gls{rf}, and \gls{resnet} models. For deep learning models, \gls{resnet} and \gls{fcnn}, features such as ``Coronary atherosclerosis of native coronary artery" and``Coronary artery disease" appear as highly important. Markers such as \gls{mchc}, \gls{mch}, neutrophils, platelet count, and eosinophils feature prominently across models. Potassium-related features show importance in several models, particularly in \gls{xgboost}, \gls{resnet}, and \gls{rf}. In models such as \gls{resnet} and \gls{fcnn}, we see several features related to more specific conditions, like unspecified essential hypertension and other hyperlipidemia. Features related to chronic conditions, like unspecified essential hypertension, acute kidney failure, and diabetes with renal manifestations, are consistently seen in models like \gls{lr}, \gls{resnet}, and \gls{fcnn}.

\subsection{Cox proportional hazards model results}

\paragraph{\gls{c-index}.} \autoref{tab:cox_model_cindex} provides the results for \gls{c-index} obtained using the \glspl{cphm}. The \gls{lr}-augmented Cox model performs the best, with an average of 0.8900 and a best score of 0.9016. Other models, including \gls{xgboost}-Augmented Cox and \gls{resnet}-Augmented Cox, show competitive results, both achieving an average of around 0.8876 and 0.8878, respectively. The baseline Cox model, with an average \gls{c-index} of 0.8820, is outperformed by all the augmented models. \gls{dt}-Augmented Cox and \gls{rf}-Augmented Cox models show slightly lower averages at 0.8855 and 0.8865, respectively.

\paragraph{Brier score.}

The \gls{cphm} Brier score results are plotted at 5 years in Appendix \ref{apd:brier}, and reported at annual intervals in \autoref{tab:cox_model_brier}. The \gls{xgboost}-augmented Cox model outperforms the other models across most time intervals, particularly at 1 year (0.0289), 4 years (0.0750), and 5 years (0.0801). \gls{fcnn}-Augmented Cox achieves the best Brier scores at 2 years (0.0485) and 3 years (0.0625). \gls{dt}-Augmented Cox also shows competitive performance, especially at 2 years with a score of 0.0496. Baseline Cox shows higher Brier scores across the intervals, especially at 5 years (0.1120), indicating less accurate predictions compared to the augmented models.

\paragraph{Dynamic \gls{auroc}.}
The \gls{cphm} dynamic \gls{auroc} results are plotted at 5 years in Appendix \ref{apd:time_dep_auc}, and reported at annual intervals in \autoref{tab:cox_model_auc}. The \gls{lr}-Augmented Cox model performs best for the first 3 years, with the highest \gls{auroc} of 0.9634 at 1 year, 0.9499 at 2 years, and 0.9453 at 3 years. However, at the 4- and 5-year intervals, \gls{xgboost}-Augmented Cox achieves the highest \gls{auroc} of 0.9376 and 0.9507, respectively. Baseline Cox performs consistently, but its scores decline slightly over time, especially at 5 years (0.9113). Other augmented models, such as \gls{fcnn} and \gls{resnet}, show stable but slightly lower AUC scores compared to the best-performing models.

\subsection{Discussion}
% The results of the experiments demonstrate that combining machine learning models with \gls{shap} value analysis can potentially identify novel predictors (in addition to those found in \gls{kfre}-8) for \gls{ckd} progression. The model performance, as measured by \gls{auroc}, highlights \gls{xgboost} as the top-performing model for feature extraction, achieving an average \gls{auroc} of 0.7796. Deep learning models such as \gls{resnet} and \gls{fcnn}, while performing slightly lower in \gls{auroc}, still contribute meaningful insights, particularly when combined with \gls{shap} values to interpret the importance of features.

Across all models, \gls{shap} analysis reveals that traditional kidney function markers, particularly creatinine (e.g., mean, max, last, and median values) and renal dialysis status, are consistently the most important features, affirming their clinical relevance in \gls{ckd} progression. Additionally, urea nitrogen (max and mean) is another key renal marker that ranks highly, emphasizing its significance in predicting disease outcomes. These results indicate that the models correctly prioritize well-established \gls{ckd} markers, providing confidence in the feature selection process.

\Gls{shap} analysis also highlights several non-traditional features that are not part of the \gls{kfre}-8 model but rank highly across models. Markers such as \gls{mchc}, \gls{mch}, neutrophils, and platelet count — blood-related features not typically associated with \gls{ckd} — emerge as important predictors. Their relevance suggests that systemic factors related to hematologic and immune responses may play a role in CKD progression. Though it is worth noting that this could also be a result of the patient population used in this study. Furthermore, potassium levels appear highly in models such as \gls{xgboost} and \gls{resnet}, reflecting the importance of this for \gls{ckd} progression prediction. Additionally, deep learning models such as \gls{resnet} and \gls{fcnn} bring attention to cardiac-related conditions, including coronary atherosclerosis and coronary artery disease, which are identified as significant predictors. This suggests that cardiovascular conditions, often comorbid with \gls{ckd}, should be considered as potential predictors in models for \gls{ckd} progression. These findings underscore the value of machine learning in surfacing predictors that may extend beyond traditional renal markers, contributing to a more comprehensive understanding of \gls{ckd}.

The application of \gls{shap} values enables not only the identification of important features but also their interpretability, allowing for the validation of established predictors while suggesting new avenues for research. By identifying features not present in the \gls{kfre}-8 model, such as creatinine levels, urea nitrogen, potassium, and blood-related markers, this analysis highlights the potential for integrating additional predictors into CKD progression models. This could improve the predictive accuracy and provide deeper insights into disease mechanisms.

%Overall, these results suggest that machine learning models, coupled with SHAP analysis, offer a powerful approach to refining and expanding the predictive models for CKD progression.% uncovering novel predictors that can enhance patient stratification and risk management.

%% file: 7-limitations.tex
\section{Limitations}
\label{sec:limitations}

The identification of novel predictors requires further validation, as these features may serve as proxies for unmodeled processes. Moreover, the reliance on the MIMIC-IV dataset, which is derived from an emergency room setting, limits the generalizability of our findings to patient cohorts within this environment. To strengthen the robustness of our results, validation using an external cohort is necessary to confirm the findings and improve the accuracy of the method.

%% file: 8-conclusion.tex
\section{Conclusion}
In this study, we propose and validate a novel approach for predicting \gls{ckd} progression by integrating machine learning models with classical statistical techniques. This method identifies potential novel predictors of CKD progression while confirming established risk factors. Our results show that combining machine learning-based feature selection with Cox proportional hazards models enhances predictive performance for CKD. 
% In this study, we develop and validate a novel approach to predicting the progression of chronic kidney disease by integrating XGboost, FCNN, and ResNet machine learning models with classical statistical techniques. Leveraging the strengths of both domains, our study intends to identify novel predictors of CKD progression that can be used to enhance early detection and treatment intervention strategies. The results demonstrate that the integration ML-based feature selection with Cox proportional hazards models not only confirmed several well-established predictors of CKD progression, but also unveiled novel predictors that have not been extensively documented in the literature. Moreover, this integration led to an improvement in predictive performance and stability in comparison a baseline model relying exclusively on traditional clinical risk factors.

% The method proposed in this study contributes to the growing body of evidence supporting the application of ML in healthcare and chronic disease management. By demonstrating the utility of a combined ML-statistical approach, this research offers a framework for the development of more robust and interpretable predictive models that can be generalized to chronic conditions beyond CKD, contributing to improved patient healthcare outcomes and reduced healthcare costs.

%% file: 9-appendix-A-figure.tex
\section{Pipeline Schematic}\label{apd:pipeline_schematic}

\begin{figure}[htbp]
    \floatconts
    {fig:ml_feature_selection} % Label goes here
    {\caption{High-level overview of the model pipeline, including ML feature selection and augmented Cox proportional hazards models.}}
    {\includegraphics[width=0.8\linewidth]{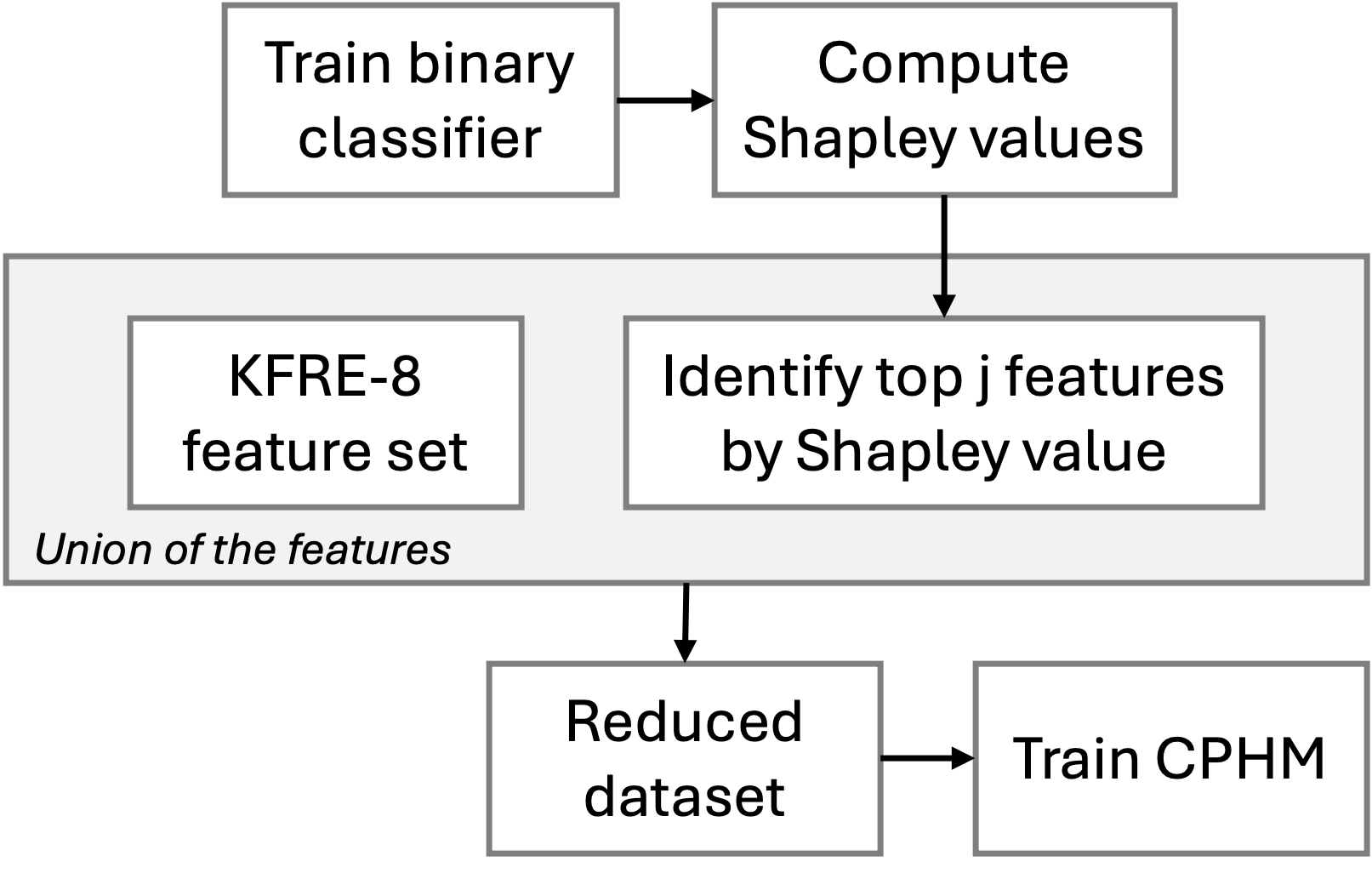}}
\end{figure}

\pagebreak

%% file: 9-appendix-B-related-work.tex
\section{Related Work}\label{apd:related}

Numerous studies have focused on predicting the outcomes of \gls{ckd} and its progression. The existing literature documents a variety of methods for developing predictive models, and identification of novel predictors yielding reasonable levels of accuracy, sensitivity, and specificity.

\citet{xiao2019comparison} assessed the utility of \gls{ml} models, including logistic regression, Elastic Net, and ensemble methods, to predict 24-hour urinary protein outcomes in CKD patients with proteinuria. Logistic regression emerged as the top performer with an AUC of 0.873, followed closely by linear models such as Elastic Net, lasso, and ridge regression. \citet{bai2022machine} compared logistic regression, naïve Bayes, random forest, decision tree, and K-nearest neighbors for predicting \gls{esrd} over five years. Random forest achieved the best AUC of 0.81, outperforming the \gls{kfre} in sensitivity.

Classical statistical techniques have also been applied to conduct nuanced survival analysis, as demonstrated in the methodology employed by \citet{ye2021nomogram}. \citet{ye2021nomogram} utilized Cox proportional hazards regression to develop a nomogram predicting three-year adverse outcomes for East Asian \gls{ckd} patients, achieving high C-statistics across datasets. The nomogram's utility was demonstrated via \gls{dca}, showing a higher net benefit compared to using \gls{egfr} alone, especially near threshold probabilities where clinical decisions are highly critical.

Leveraging a combined methodology integrating both ML for enhanced feature selection and classical statistical techniques for survival analysis, \citet{liu2023combining} innovated an improved pipeline for predicting breast cancer in post-menopausal women. The method began with the use of the extreme gradient boosting (XGBoost) algorithm to identify novel predictors from over 1,700 features recorded in the UK Biobank, ranking feature importance and highlighting key relationships with Shapley Additive exPlanations (SHAP) values. The integration of ML in this feature selection process was intended as a strategy to enhance identification of non-linear relationships among predictor features and effectively manage the high-dimensional data. Following feature selection, the researchers constructed two Cox proportional hazards models: a baseline model incorporating known risk factors and polygenic risk scores (PRS), and an augmented model including an additional set of ten novel predictor features identified by XGBoost. The findings demonstrated that the augmented Cox model showed improved predictive performance compared to the baseline model, achieving higher Harrell’s C-index scores.

% Our pipeline aims to adapt and optimize  \citet{liu2023combining}'s method, applying it to develop a predictive model that can aid in early detection and patient management in \gls{ckd} treatment.

% Short paragraph to refer to additional reading, refer to Ammar thesis for stuctural reference

%% file: 9-appendix-B1-kfre.tex
\section{Kidney Failure Risk Equation}\label{apd:kfre}
The \gls{kfre} is a predictive tool used to estimate the likelihood of a \gls{ckd} patient progressing to \gls{esrd} typically within a 2 to 5 year time frame \citep{tangri2016multinational}. These equations have been validated across various populations globally, and have shown to effectively aid clinical decision-making, particularly with regard to determining the timing of treatment interventions such as dialysis or transplantation \citep{major2019kfre}.

The \Gls{kfre} is available in both 4-variable (\gls{kfre}-4) and 8-variable model (\gls{kfre}-8) versions \citep{nkf2023kfre}. \Gls{kfre}-4 makes use of age, sex, \gls{egfr}, and \gls{uacr}. \gls{kfre}-8 includes these same factors as \gls{kfre}-4 with the addition of serum calcium, serum phosphorus, serum bicarbonate, and serum albumin.

%% file: 9-appendix-C-binary-classifier.tex
\section{Binary Classifier}\label{apd:binary_class}

Let $\tau \in \mathbb{R}$ be a threshold value, and let $f(\cdot;\mathbf{\theta}):\mathbb{R}^{n} \mapsto \{0,1\}$ represent a function parameterized by $\mathbf{\theta}$. For a data sample $\mathbf{x}_{i} \in \mathbb{R}^{n}$, a binary classifier assigns a label $\hat{y}_{i}$ based on $\tau$ as,
\begin{equation}
\label{eqn_binary_classifier}
    \hat{y}_i = 
\begin{cases} 
0, & f(\mathbf{x}_i; \mathbf{\theta}) < \tau, \\
1, & f(\mathbf{x}_i; \mathbf{\theta}) \geq \tau.
\end{cases}
\end{equation}
The function $f(\cdot; \mathbf{\theta})$ can be represented by a variety of machine learning models. In this paper, we focus on tree-based methods \ref{apd:tree_methods}, linear models \ref{apd:linear_models}, and deep learning approaches \ref{apd:neural_networks}, each capable of performing binary classification as defined in equation (\ref{eqn_binary_classifier}).

\subsection{Linear models}\label{apd:linear_models}
\label{sec:linear_models}
\paragraph{Logistic regression.} Logistic regression is a fundamental statistical modeling technique used to estimate the probability of a binary outcome, based on one or more predictor variables \citep{hosmer2013applied}. Let \( \mathbf{x}_i \in \mathbb{R}^{n_0} \) represent input features and \( \mathbf{w} \in \mathbb{R}^{n_0} \) denote a vector of coefficients to be learned. The probability of the binary event occurring \( y_i \in \{0,1\} \) for the \( i \)-th data sample, is defined as
\begin{equation}
\label{logreg_prediction}
\hat{y}_i = \sigma(\mathbf{w}^T \mathbf{x}_i + b).
\end{equation}
% The model parameters \( \mathbf{w} \) and \( b \) are updated using gradient descent to iteratively reduce the loss, making logistic regression a suitable choice for approximating a decision boundary for binary classification tasks such as those discussed in \eqref{eqn_binary_classifier}.

\subsection{Tree-based methods} \label{apd:tree_methods}
\label{sec:tree_based}
\paragraph{Decision tree.} A decision tree is a non-parametric supervised learning algorithm that recursively partitions the input space based on feature values, forming a tree structure composed of decision nodes and leaf nodes, where each leaf node assigns a predicted class label \citep{breiman1984classification}. The prediction for a data point is given by
\begin{equation}
\label{decision_tree_prediction}
\hat{y}_i = \sum_{l=1}^{L} p_l \mathbb{I}\{ \mathbf{x}_i \in R_l \}
\end{equation}
where \( p_l \) is the predicted class probability for region \( R_l \), and \( \mathbb{I}\{ \mathbf{x}_i \in R_l \} \) is an indicator function that equals 1 if the sample \( \mathbf{x}_i \) falls into region \( R_l \), and 0 otherwise.

%Decision trees can be applied to approximate the decision boundary for binary classification tasks \eqref{eqn_binary_classifier}.

\paragraph{Random forest.} A random forest is an ensemble learning method, aggregating multiple decision trees leads to improve model robustness and generalization \citep{breiman2001random}. For binary classification, the prediction \( \hat{y}_i \) for a given data point \( \mathbf{x}_i \in \mathbb{R}^{n_0} \) is obtained by aggregating the predictions from \( T \) individual decision trees in the forest
\begin{equation}
\label{random_forest_prediction}
\hat{y}_i = \frac{1}{T} \sum_{t=1}^{T} \hat{y}_i^{(t)}
\end{equation}
Here, \( \hat{y}_i^{(t)} \) denotes the prediction from the \( t \)-th decision tree. %Each tree is trained on a different bootstrap sample of the data, and during training, a random subset of features is considered at each split, introducing diversity among the trees and reducing the risk of overfitting.

%The random forest model can be applied to approximate the decision boundary for binary classification tasks \eqref{eqn_binary_classifier}.

\paragraph{Extreme gradient boosting.} Extreme Gradient Boosting (XGBoost) is a scalable and efficient \gls{ml} algorithm, effectively enhancing model accuracy through the combination of multiple weak learners \citet{chen2016xgboost}. The prediction for a data sample \( i \) can be expressed as the weighted sum of the outputs from \( K \) decision trees
\begin{equation}
\label{eqn_xgboost}
\hat{y}_i = \sum_{k=1}^{K} f_k(\mathbf{x}_i)
\end{equation}
where \( f_k \) denotes the \( k \)-th tree's contribution, and \( \mathbf{x}_i \in \mathbb{R}^{n_0} \) represents the input features of sample \( i \). The ensemble of trees allows the model to capture complex non-linear patterns, much like the layers in a neural network.

% The XGBoost objective function integrates a loss component and a regularization component to prevent overfitting
% \begin{equation}
% \label{eqn_xgboost_objective}
% L(\Theta) = \sum_{i=1}^{n} l(\hat{y}_i, y_i) + \sum_{k=1}^{K} \Omega(f_k)
% \end{equation}
% Here, \( l(\hat{y}_i, y_i) \) is a differentiable loss function quantifying the error for each prediction, and \( \Omega(f_k) \) is a regularization term that penalizes model complexity, including aspects such as tree depth and leaf count. By minimizing this objective function through gradient descent update rules similar to \eqref{update_rule}, XGBoost can be applied to approximate a function \( f(\mathbf{x}_i; \mathbf{\theta}) \) for used in binary classification \eqref{eqn_binary_classifier}.

\subsection{Neural networks} \label{apd:neural_networks}

\paragraph{Fully-connected neural network.} A fully connected neural network can be defined as
\begin{equation}
\label{fcnn}
\small
\hat{y}_i = \sigma(W_L (\sigma(W_{L-1} (\cdots \sigma(W_1 \mathbf{x}_i + \mathbf{b}_1) \cdots) + \mathbf{b}_{L-1}) + \mathbf{b}_L))
\end{equation}
where \( W_l \in \mathbb{R}^{n_l \times n_{l-1}} \) and \( \mathbf{b}_l \in \mathbb{R}^{n_l} \) represent the weights and biases for each layer \( l \in \{1, 2, \dots, L\} \), and \( \sigma \) is an activation function such as the \gls{relu} \citep{lecun2015deep}. The input vector \( \mathbf{x}_i \in \mathbb{R}^{n_0} \) represents the features for data sample \( i \), and the output \( \hat{y}_i \) is the predicted label. The activation function \( \sigma \) introduces non-linearity, enabling the network to model complex patterns.

\paragraph{Residual neural network.} A \gls{resnet} addresses the degradation problem in excessively deep networks, wherein added depth increases training error, through the introduction of shortcut connections \citep{he2016resnet}. The output of a residual block can be defined as
\begin{equation}
\label{resblock}
\mathbf{y}_i = F(\mathbf{x}_i, \{W_i\}) + \mathbf{x}_i
\end{equation}
where \( \mathbf{x}_i \in \mathbb{R}^{n_0} \) is the input vector for data sample \( i \), and \( \mathbf{y}_i \) is the output of the residual block. The function \( F \) consists of weights \( \{W_i\} \) where \( W_i \in \mathbb{R}^{n_{l} \times n_{l-1}} \), as well as the biases and activation functions for each layer \( l \) across the respective layers within the block \eqref{fcnn}, representing the residual mapping. The addition of the input \( \mathbf{x}_i \) to the output of \( F \) characterizes a shortcut connection, allowing gradients to flow directly through the network with no additional computational complexity or training error degradation.

To align the dimensions of \( \mathbf{x}_i \) with those of \( F(\mathbf{x}_i, \{W_i\}) \), a linear transformation \( W_s \) may be applied to \( \mathbf{x}_i \)
\begin{equation}
\label{resblock_modified}
\mathbf{y}_i = F(\mathbf{x}_i, \{W_i\}) + W_s \mathbf{x}_i
\end{equation}
where \( W_s \in \mathbb{R}^{n_l \times n_0} \). %The network is trained with the gradient descent update rule \eqref{update_rule}, and used to approximate a function \( f(\mathbf{x}_i; \mathbf{\theta}) \) as part of the binary classification task \eqref{eqn_binary_classifier}.

%% file: 9-appendix-C2-shap.tex
\section{Shapley values}\label{apd:shap}

Shapley values are a concept from cooperative game theory providing a fair way to distribute total gains or cost amongst multiple players in a coaliton \citep{shapley1953value}. Let $M=\{1,..,m\}$ represent a set of players taking part in a game $\nu : 2^{M} \mapsto \mathbb{R}$. The game, $\nu$ is a characteristic function returning a scalar reward for each coalition $A \subseteq M$. It is assumed that $\nu(\emptyset)=0$. The Shapley value of a player $k$ can be computed by
\begin{equation}
\label{eqn_shapley_values}
    \phi_{k} = \Sigma_{A \subseteq M \setminus \{k\}} \frac{(m-|A|-1)!|A|!}{m!}(\nu(A \cup \{k\})-\nu(A)).
\end{equation}
Here $|\cdot|$ represents cardinality. Shapley values are commonly used in explainable machine learning \citep{merrick2020explanation}. In the context of machine learning, each feature $x_{i}$ in a feature vector $\mathbf{x} \in \mathbb{R}^{n}$ is considered as the player. While the output from a trained model $f:\mathbb{R}^{n} \mapsto \mathbb{R}$ is considered to be reward $\nu$. As a result, $\phi_{i}$ measures the $x_{i}$-th feature's importance with respect to model output.

%% file: 9-appendix-C3-cphm.tex
\section{Cox proportional hazards model}\label{apd:cphm}

The \glsfirst{cphm} is a statistical technique used for modeling the time-dependent risk of an event occurring, often applied in survival analysis \citep{cox1972regression}. Let $\mathbf{x} \in \mathbb{R}^{d}$ represent a feature vector of $d$-dimension and let $\mathbf{\beta} \in \mathbb{R}^{d}$ denote a corresponding vector of coefficients. The \gls{cphm} is based on a hazard function, defined by
\begin{equation}
\lambda(t \mid x) = \lambda_0(t) \exp(\beta^T x).
\label{eq:cox_not}
\end{equation}
Here $\lambda_0(\cdot): \mathbb{R}^{+} \mapsto \mathbb{R}^{+}$ is the baseline hazard function, representing the hazard rate when $\mathbf{x}=0$. The baseline hazard function $\lambda_0(\cdot)$ is unspecified, allowing for a semi-parametric approach where the effect of the $\mathbf{x}$ is modeled parametrically through $\beta$, while the baseline hazard can vary non-parametrically over time. This flexibility makes the \gls{cphm} a powerful tool for survival analysis, as it does not assume a specific distribution for the survival times.

\paragraph*{Partial likelihood.}
The parameters \( \beta \) are estimated using the method of partial likelihood, which focuses on the order of events rather than their exact timing. The partial likelihood function \( L(\beta) \) is given by
\begin{equation}
L(\beta) = \prod_{i=1}^D \frac{\exp(\beta^T x_i)}{\sum_{j \in R(t_i)} \exp(\beta^T x_j)},
\label{eq:part_like}
\end{equation}
where \( D \) is the number of observed events, and \( R(t_i) \) is the risk set at time \( t_i \), consisting of all individuals who are at risk just prior to time \( t_i \). The partial likelihood function is maximized to obtain the estimates of \( \beta \).

%% file: 9-appendix-D-eval-metrics.tex
\section{Evaluation Metrics}\label{apd:eval_metric}

\subsection{Area under the receiver operating characteristic curve.}\label{subsec:auroc} The \gls{auroc} is a metric for evaluating the performance of binary classification models. It measures the ability of a model to distinguish between positive and negative classes by plotting the true positive rate against the false positive rate across various decision thresholds. \gls{auroc} score ranges from 0 to 1, where 1 indicates perfect discrimination, 0.5 represents random guessing. In survival analysis, a time-dependent version of the \gls{auroc} is often applied to evaluate predictive accuracy at different time points.

\subsection{Concordance index.}\label{subsec:cindex} The \gls{c-index} is a metric for evaluating the predictive accuracy of survival models, assessing how well the predicted and actual order of events agree \citep{harrell1982evaluating}. Let $T$ denote the event time, \gls{c-index} is then defined by,
\begin{equation}
    \mathcal{C}=\frac{\sum_{i,j} \mathds{1}[T_{i} < T_{j}] \mathds{1}[\hat{T}_{i} < \hat{T}_{j}] \delta_{i}}{\sum_{i,j} \mathds{1}[T_{i} < T_{j}] \delta_{i}}.
\end{equation}
Here, $\delta_{i} = 1$ if the event time $T_{i}$ is observed (i.e. not censored), and $\delta_{i} = 0$ otherwise. \gls{c-index} is a generalisation of the \gls{auroc}. A \gls{c-index} of 1 indicates perfect predictions and 0 represents the worst performance.

\subsection{Brier score.}\label{subsec:brier} The Brier score is a metric used to evaluate the accuracy of probabilistic predictions in survival analysis \citep{brier1950verification}. It measures the weighted mean squared difference between the predicted probabilities and the actual outcomes. Brier score at time $t$ is defined by,
\begin{equation}
    \mathcal{B}(t) = \frac{1}{N} \sum_{i=1}^{N} \left( \mathds{1}[T_{i} > t] - \hat{S}(t | \mathbf{x}_{i})\right)^{2},
\end{equation}
where $\hat{S}(\cdot)$ represents a predicted survival function. The Brier score ranges from 0 to 1, with 0 indicating perfect accuracy and 1 representing the worst accuracy.
% https://square.github.io/pysurvival/metrics/c_index.html

%% file: 9-appendix-E-supp-tables.tex
\section{Supplementary Tables}\label{apd:table_supplement}

\subsection{Data characteristics}\label{apd:data_tabs}

\begin{table}[htbp]
\floatconts
  {tab:ckd_icd9} % Label for referencing the table
  {\caption{CKD-related ICD-9 codes in MIMIC-IV} \label{tab:ckd_icd9}} % Correct placement of the label
  {\centering % Center the table contents
  \resizebox{\columnwidth}{!}{ % Resize table to fit within the column
  \begin{tabular}{l l}
    \toprule
    \textbf{ICD-9 Code} & \textbf{Long Title} \\
    \midrule
    \texttt{5851} & Chronic kidney disease, Stage I \\
    \texttt{5852} & Chronic kidney disease, Stage II (mild) \\
    \texttt{5853} & Chronic kidney disease, Stage III (moderate) \\
    \texttt{5854} & Chronic kidney disease, Stage IV (severe) \\
    \texttt{5855} & Chronic kidney disease, Stage V \\
    \texttt{5856} & End stage renal disease \\
    \texttt{5859} & Chronic kidney disease, unspecified \\
    \bottomrule
  \end{tabular}
  }
  }
\end{table}

\begin{table}[htbp]
\floatconts
  {tab:ml_in} % Label for the table
  {\caption{Input categories of the 1,373 input features used in ML feature selection.} \label{tab:ml_in}} % Caption
  {\centering % Center the table contents
  \resizebox{0.8\columnwidth}{!}{ % Resize table to fit within the column
  \begin{tabular}{l l}
    \toprule
      \textbf{Input Category} & \textbf{Frequency} \\
      \midrule
      Patient Demographics & 6 (0.44\%) \\
      CKD Diagnostic Information & 3 (0.22\%) \\
      Clinical Lab Data & 444 (32.34\%) \\
      Comorbid Health Conditions & 921 (67.08\%) \\
      \bottomrule
  \end{tabular}
  }
  }
\end{table}

\subsection{Hyperparameters}\label{apd:hparams}

\begin{table}[htbp]
\floatconts
  {tab:nn_hparams} % Label for referencing the table
  {\caption{Manually selected hyperparameter values for \gls{fcnn} and \gls{resnet}.}} % Caption for the table
  {\centering % Center the table contents
  \resizebox{0.7\columnwidth}{!}{ % Resize table to fit within the column
  \begin{tabular}{l c c}
    \toprule
    \textbf{Hyperparameter} & \textbf{FCNN} & \textbf{ResNet} \\
    \midrule
    \texttt{learning\_rate}     & $0.001$ & $0.001$ \\
    \texttt{weight\_decay}      & $1 \times 10^{-4}$ & $1 \times 10^{-4}$ \\
    \texttt{max\_epochs}        & $35$ & $30$ \\
    \texttt{early\_stopping}           & $8$ & $5$ \\
    \bottomrule
  \end{tabular}
  }
  }
\end{table}

\pagebreak

\begin{table}[htbp]
\floatconts
  {tab:bayes} % Label for referencing the table
  {\caption{Optimized hyperparameter values and search spaces for Bayesian optimization.}} % Caption for the table
  {\centering % Center the table contents
  \resizebox{\columnwidth}{!}{ % Resize table to fit within the column
  \begin{tabular}{l c c c c}
    \toprule
    \textbf{Hyperparameter} & \textbf{XGBoost} & \textbf{RF} & \textbf{DT} & \textbf{LR} \\
    \midrule
    \texttt{max\_depth}        & $14\ (5-20)$  & $33\ (2-50)$ & $6\ (1-50)$ & \emph{NA} \\
    \texttt{min\_samples\_split}      & \emph{NA}  & $6\ (2-20)$  & $6\ (2-20)$ & \emph{NA} \\
    \texttt{min\_samples\_leaf}       & \emph{NA}  & $7\ (1-20)$  & $11\ (1-20)$ & \emph{NA} \\
    \texttt{n\_estimators}        & $83\ (50-150)$  & $138\ (50-200)$ & \emph{NA} & \emph{NA} \\
    \texttt{min\_child\_weight}        & $1\ (1-10)$  & \emph{NA} & \emph{NA} & \emph{NA} \\
    \texttt{gamma}        & $2.61\ (0.5-3.0)$  & \emph{NA} & \emph{NA} & \emph{NA} \\
    \texttt{subsample}        & $0.78\ (0.6-1.0)$  & \emph{NA} & \emph{NA} & \emph{NA} \\
    \texttt{colsample\_bytree}        & $0.76\ (0.6-1.0)$  & \emph{NA} & \emph{NA} & \emph{NA} \\
    \texttt{colsample\_bylevel}        & $0.56\ (0.01-0.6)$  & \emph{NA} & \emph{NA} & \emph{NA} \\
    \texttt{learning\_rate}        & $0.22\ (0.01-0.3)$  & \emph{NA} & \emph{NA} & \emph{NA} \\
    \texttt{C}        & \emph{NA}  & \emph{NA} & \emph{NA} & $0.095\ (0.01-1.0)$ \\
    \bottomrule
  \end{tabular}
  }
  }
\end{table}

\subsection{Model performance}\label{apd:model-performace}

\begin{table}[htbp]
\floatconts
  {tab:ml_model_auc} % Label for the table
  {\caption{\gls{auroc} scores for models used for feature extraction.}} % Caption
  {\centering
  \resizebox{\columnwidth}{!}{ % Resizes the table to fit within the column width
  \begin{tabular}{l c c}
  \toprule
  \textbf{Model} & \textbf{AUC-ROC Avg.} & \textbf{AUC-ROC Best} \\
  \midrule
  XGBoost & $\mathbf{0.7796\pm0.0257}$  & $0.8105$ \\
  LR & $0.7027\pm0.0230$  & $0.7268$ \\
  DT & $0.7283\pm0.0364$  & $0.7799$ \\
  RF & $0.5796\pm0.0092$  & $0.5905$ \\
  FCNN  & $0.6612\pm0.0179$  & $0.6916$ \\
  ResNet & $0.6540\pm0.0090$  & $0.6661$ \\
  \bottomrule
  \end{tabular}
  }}
\end{table}

\begin{table}[htbp]
\floatconts
  {tab:cox_model_cindex} % Label for referencing the table
  {\caption{Comparison of Cox proportional hazards models based on C-Index scores obtained from five-fold cross-validation.}} % Caption for the table
  {\centering % Center the table contents
  \resizebox{\columnwidth}{!}{ % Resize table to fit within the column
  \small
  \begin{tabular}{l c c}
    \toprule
    \textbf{Model} & \textbf{C-Index Avg.} & \textbf{C-Index Best} \\
    \midrule
    Baseline Cox        & $0.8820\pm0.0082$  & $0.8910$ \\
    XGBoost-Augmented Cox & $0.8876\pm0.0071$  & $0.8929$ \\
    LR-Augmented Cox & $\mathbf{0.8900\pm0.0092}$  & $0.9016$ \\
    DT-Augmented Cox & $0.8855\pm0.0051$  & $0.8901$ \\
    RF-Augmented Cox & $0.8865\pm0.0082$  & $0.8936$ \\
    FCNN-Augmented Cox    & $0.8873\pm0.0073$  & $0.8942$ \\
    ResNet-Augmented Cox  & $0.8878\pm0.0073$  & $0.8930$ \\
    \bottomrule
  \end{tabular}
  }
  }
\end{table}

\begin{table}[htbp]
\floatconts
  {tab:cox_model_brier} % Label for referencing the table
  {\caption{Comparison of Cox proportional hazards models based on Brier scores at 1-5 year time intervals.}} % Caption for the table
  {\centering % Center the table contents
  \resizebox{\columnwidth}{!}{ % Resize table to fit within the column
  \begin{tabular}{l c c c c c}
    \toprule
    \textbf{Model} & \textbf{1 Year} & \textbf{2 Year} & \textbf{3 Year} & \textbf{4 Year} & \textbf{5 Year} \\
    \midrule
    Baseline Cox        & $0.0321$  & $0.0581$ & $0.0730$ & $0.0901$ & $0.1120$ \\
    XGBoost-Augmented Cox & $\mathbf{0.0289}$  & $0.0506$ & $0.0673$ & $\mathbf{0.0750}$ & $\mathbf{0.0801}$ \\
    LR-Augmented Cox & $0.0308$  & $0.0550$ & $0.0673$ & $0.0830$ & $0.1033$ \\
    DT-Augmented Cox & $0.0309$  & $0.0496$ & $0.0646$ & $0.0841$ & $0.1010$ \\
    RF-Augmented Cox & $0.0321$  & $0.0590$ & $0.0732$ & $0.0899$ & $0.1077$ \\
    FCNN-Augmented Cox    & $0.0299$  & $\mathbf{0.0485}$ & $\mathbf{0.0625}$ & $0.0828$ & $0.0988$ \\
    ResNet-Augmented Cox  & $0.0321$  & $0.0572$ & $0.0710$ & $0.0852$ & $0.1079$ \\
    \bottomrule
  \end{tabular}
  }
  }
\end{table}

\begin{table}[htbp]
\floatconts
  {tab:cox_model_auc} % Label for referencing the table
  {\caption{Comparison of Cox proportional hazards models based on AUC scores at 1-5 year time intervals.}} % Caption for the table
  {\centering % Center the table contents
  \resizebox{\columnwidth}{!}{ % Resize table to fit within the column
  \begin{tabular}{l c c c c c}
    \toprule
    \textbf{Model} & \textbf{1 Year} & \textbf{2 Year} & \textbf{3 Year} & \textbf{4 Year} & \textbf{5 Year} \\
    \midrule
    Baseline Cox        & $0.9551$  & $0.9369$ & $0.9357$ & $0.9212$ & $0.9113$ \\
    XGBoost-Augmented Cox & $0.9494$  & $0.9171$ & $0.9071$ & $\mathbf{0.9376}$ & $\mathbf{0.9507}$ \\
    LR-Augmented Cox & $\mathbf{0.9634}$  & $\mathbf{0.9499}$ & $\mathbf{0.9453}$ & $0.9300$ & $0.9234$ \\
    DT-Augmented Cox & $0.9340$  & $0.9494$ & $0.9407$ & $0.9090$ & $0.9111$ \\
    RF-Augmented Cox & $0.9472$  & $0.9426$ & $0.9399$ & $0.9212$ & $0.9149$ \\
    FCNN-Augmented Cox    & $0.9364$  & $0.9466$ & $0.9422$ & $0.9169$ & $0.9149$ \\
    ResNet-Augmented Cox  & $0.9449$  & $0.9423$ & $0.9419$ & $0.9266$ & $0.9167$ \\
    \bottomrule
  \end{tabular}
  }
  }
\end{table}

\pagebreak

\subsection{Feature selection}\label{apd:model-performace}

\begin{table}[htbp]
\floatconts
  {tab:cox_sum_xgb}
  {\caption{Hazard ratios, 95\% confidence intervals (CI), and p-values for XGBoost-augmented Cox model.}}
  {\centering
  \resizebox{\columnwidth}{!}{
  % \small
  \begin{tabular}{l c c}
    \toprule
    \textbf{Features} & \textbf{Hazard Ratio (95\% CI)} & \textbf{p-value} \\
    \midrule
    \texttt{Creatinine\_mean}        & $1.0995\ (0.9403,\ 1.2856)$  & $0.2345$ \\
    \texttt{Creatinine\_max}         & $0.9807\ (0.9436,\ 1.0193)$  & $0.3228$ \\
    \texttt{Renal dialysis status}   & $1.7696\ (1.4469,\ 2.1644)$  & $2.77 \times 10^{-8}$ \\
    \texttt{Creatinine\_last}        & $1.0287\ (0.9807,\ 1.0790)$  & $0.2455$ \\
    \texttt{Creatinine\_median}      & $0.8753\ (0.7823,\ 0.9794)$  & $0.0202$ \\
    \texttt{Urea\_Nitrogen\_max}     & $1.0031\ (1.0002,\ 1.0060)$  & $0.0371$ \\
    \texttt{Potassium\_max}          & $1.0790\ (0.8686,\ 1.3403)$  & $0.4920$ \\
    \texttt{Creatinine\_Urine\_min}  & $0.9979\ (0.9958,\ 1.0001)$  & $0.0644$ \\
    \texttt{Urea\_Nitrogen\_last}    & $0.9978\ (0.9945,\ 1.0011)$  & $0.1864$ \\
    \texttt{Unspecified essential hypertension} & $0.5960\ (0.5211,\ 0.6817)$  & $4.28 \times 10^{-14}$ \\
    \texttt{Orthostatic hypotension} & $1.1601\ (0.9631,\ 1.3976)$  & $0.1179$ \\
    \texttt{White\_Blood\_Cells\_min} & $0.9822\ (0.9446,\ 1.0212)$  & $0.3655$ \\
    \texttt{Bilirubin\_Total\_mean}  & $1.0254\ (0.9757,\ 1.0777)$  & $0.3219$ \\
    \texttt{Protein/Creatinine\_Ratio\_max} & $0.9892\ (0.9804,\ 0.9980)$  & $0.0168$ \\
    \texttt{Neutrophils\_max}        & $0.9964\ (0.9882,\ 1.0046)$  & $0.3833$ \\
    \texttt{Platelet\_Count\_median} & $1.0007\ (0.9994,\ 1.0020)$  & $0.2853$ \\
    \texttt{Hemoglobin\_max\_y}      & $0.8836\ (0.8473,\ 0.9214)$  & $7.34 \times 10^{-9}$ \\
    \texttt{Eosinophils\_first}      & $1.0253\ (1.0044,\ 1.0466)$  & $0.0176$ \\
    \texttt{Cholesterol\_Total\_last} & $0.9999\ (0.9984,\ 1.0014)$  & $0.9190$ \\
    \texttt{MCHC\_min}               & $1.0058\ (0.9567,\ 1.0573)$  & $0.8222$ \\
    \texttt{Urea\_Nitrogen\_first}   & $1.0018\ (0.9977,\ 1.0058)$  & $0.3903$ \\
    \texttt{Glucose\_min}            & $0.9971\ (0.9944,\ 0.9999)$  & $0.0391$ \\
    \texttt{Creatinine\_min}         & $1.3787\ (1.2236,\ 1.5535)$  & $1.34 \times 10^{-7}$ \\
    \texttt{Uric\_Acid\_first}       & $1.0220\ (0.9897,\ 1.0553)$  & $0.1844$ \\
    \texttt{Acute kidney failure, unspecified} & $1.0080\ (0.8587,\ 1.1833)$  & $0.9226$ \\
    \texttt{Total Protein\_Urine\_last} & $1.0007\ (1.0001,\ 1.0013)$ & $0.0262$ \\
    \texttt{Mitral valve disorders}  & $0.9100\ (0.7684,\ 1.0777)$  & $0.2747$ \\
    \texttt{Potassium\_mean}         & $0.7925\ (0.5992,\ 1.0480)$  & $0.1028$ \\
    \texttt{Protein/Creatinine\_Ratio\_last} & $1.0288\ (1.0099,\ 1.0480)$  & $0.0027$ \\
    \texttt{Platelet\_Count\_max}    & $0.9994\ (0.9988,\ 1.0001)$  & $0.0960$ \\
    \texttt{MCH\_max}                & $1.0056\ (0.9814,\ 1.0304)$  & $0.6525$ \\
    \texttt{marital\_status\_MARRIED} & $0.9824\ (0.8686,\ 1.1110)$  & $0.7772$ \\
    \texttt{Sodium\_last}            & $1.0026\ (0.9874,\ 1.0181)$  & $0.7382$ \\
    \texttt{Anemia, unspecified}     & $0.9008\ (0.7920,\ 1.0245)$  & $0.1116$ \\
    \texttt{Creatinine\_Urine\_last} & $1.0006\ (0.9991,\ 1.0021)$  & $0.4252$ \\
    \texttt{Asparate\_Aminotransferase\ (AST)\_first} & $0.9993\ (0.9985,\ 1.0002)$ & $0.1381$ \\
    \texttt{Globulin\_last}          & $0.9332\ (0.8227,\ 1.0585)$  & $0.2819$ \\
    \texttt{Urea\_Nitrogen\_median}  & $1.0090\ (1.0030,\ 1.0150)$  & $0.0031$ \\
    \texttt{White\_Blood\_Cells\_mean} & $1.0057\ (0.9855,\ 1.0264)$  & $0.5813$ \\
    \texttt{Eosinophils\_max}        & $0.9933\ (0.9805,\ 1.0063)$  & $0.3133$ \\
    \bottomrule
  \end{tabular}
  }
  }
\end{table}

\begin{table}[htbp]
\floatconts
  {tab:cox_sum_fcnn}
  {\caption{Hazard ratios, 95\% confidence intervals (CI), and p-values for FCNN-augmented Cox model.}}
  {\centering
  \resizebox{\columnwidth}{!}{
  % \small
  \begin{tabular}{l c c}
    \toprule
    \textbf{Features} & \textbf{Hazard Ratio (95\% CI)} & \textbf{p-value} \\
    \midrule
    \parbox[t]{8cm}{\texttt{Renal dialysis status}} & $1.805\ (1.465,\ 2.224)$ & $2.92 \times 10^{-8}$ \\
    \parbox[t]{8cm}{\texttt{Coronary atherosclerosis \\ of native coronary artery}} & $0.972\ (0.8548,\ 1.117)$ & $0.6876$ \\
    \parbox[t]{8cm}{\texttt{Creatinine\_last}} & $1.021\ (0.9729,\ 1.071)$ & $0.4013$ \\
    \parbox[t]{8cm}{\texttt{Anemia, unspecified}} & $0.9664\ (0.8501,\ 1.099)$ & $0.6012$ \\
    \parbox[t]{8cm}{\texttt{race\_WHITE}} & $1.014\ (0.8864,\ 1.160)$ & $0.8418$ \\
    \parbox[t]{8cm}{\texttt{Other and unspecified hyperlipidemia}} & $0.9728\ (0.8472,\ 1.117)$ & $0.6952$ \\
    \parbox[t]{8cm}{\texttt{Unspecified essential hypertension}} & $0.5962\ (0.5204,\ 0.6830)$ & $8.75 \times 10^{-14}$ \\
    \parbox[t]{8cm}{\texttt{MCHC\_min}} & $1.045\ (0.9899,\ 1.102)$ & $0.1117$ \\
    \parbox[t]{8cm}{\texttt{Chloride\_max}} & $1.012\ (0.9944,\ 1.030)$ & $0.1837$ \\
    \parbox[t]{8cm}{\texttt{marital\_status\_SINGLE}} & $1.144\ (0.9956,\ 1.313)$ & $0.0577$ \\
    \parbox[t]{8cm}{\texttt{Urea\_Nitrogen\_median}} & $1.003\ (0.9897,\ 1.016)$ & $0.6907$ \\
    \parbox[t]{8cm}{\texttt{Potassium\_min}} & $0.8744\ (0.6962,\ 1.098)$ & $0.2484$ \\
    \parbox[t]{8cm}{\texttt{Urea\_Nitrogen\_mean}} & $1.013\ (0.9949,\ 1.031)$ & $0.1625$ \\
    \parbox[t]{8cm}{\texttt{Long-term use of other medications}} & $0.7729\ (0.6029,\ 0.9908)$ & $0.0420$ \\
    \parbox[t]{8cm}{\texttt{Creatinine\_min}} & $1.297\ (1.156,\ 1.455)$ & $9.91 \times 10^{-6}$ \\
    \parbox[t]{8cm}{\texttt{Urea\_Nitrogen\_last}} & $0.9984\ (0.9951,\ 1.002)$ & $0.3435$ \\
    \parbox[t]{8cm}{\texttt{Platelet\_Count\_last}} & $1.0003\ (0.9997,\ 1.0009)$ & $0.3399$ \\
    \parbox[t]{8cm}{\texttt{Acute kidney failure, unspecified}} & $1.080\ (0.9205,\ 1.267)$ & $0.3456$ \\
    \parbox[t]{8cm}{\texttt{Urea\_Nitrogen\_min}} & $1.004\ (0.9945,\ 1.014)$ & $0.4104$ \\
    \parbox[t]{8cm}{\texttt{Urea\_Nitrogen\_first}} & $1.001\ (0.9970,\ 1.0045)$ & $0.7037$ \\
    \parbox[t]{8cm}{\texttt{Personal history of TIA and cerebral \\ infarction without residual deficits}} & $0.9398\ (0.8046,\ 1.098)$ & $0.4334$ \\
    \parbox[t]{8cm}{\texttt{Urea\_Nitrogen\_max}} & $1.001\ (0.9970,\ 1.0046)$ & $0.6792$ \\
    \parbox[t]{8cm}{\texttt{Neutrophils\_min}} & $0.9928\ (0.9837,\ 1.0019)$ & $0.1194$ \\
    \parbox[t]{8cm}{\texttt{Diabetes with renal manifestations}} & $0.9806\ (0.8457,\ 1.137)$ & $0.7954$ \\
    \parbox[t]{8cm}{\texttt{MCHC\_max}} & $0.9086\ (0.8559,\ 0.9645)$ & $0.0017$ \\
    \parbox[t]{8cm}{\texttt{Peripheral vascular disease, unspecified}} & $0.9203\ (0.7872,\ 1.076)$ & $0.2974$ \\
    \parbox[t]{8cm}{\texttt{Surgical operation with transplant \\ causing abnormal reaction}} & $1.069\ (0.8494,\ 1.345)$ & $0.5694$ \\
    \parbox[t]{8cm}{\texttt{White\_Blood\_Cells\_min}} & $0.9918\ (0.9601,\ 1.024)$ & $0.6173$ \\
    \parbox[t]{8cm}{\texttt{Potassium\_max}} & $0.9548\ (0.8047,\ 1.133)$ & $0.5962$ \\
    \parbox[t]{8cm}{\texttt{Creatinine\_Urine\_last}} & $0.9994\ (0.9981,\ 1.0006)$ & $0.3338$ \\
    \parbox[t]{8cm}{\texttt{Percutaneous transluminal coronary \\ angioplasty status}} & $1.032\ (0.8793,\ 1.212)$ & $0.6968$ \\
    \parbox[t]{8cm}{\texttt{Sodium\_first}} & $1.010\ (0.9922,\ 1.028)$ & $0.2764$ \\
    \parbox[t]{8cm}{\texttt{Creatinine\_median}} & $0.9610\ (0.8802,\ 1.049)$ & $0.3753$ \\
    \parbox[t]{8cm}{\texttt{Hemoglobin\_first\_y}} & $0.9300\ (0.8966,\ 0.9647)$ & $1.03 \times 10^{-4}$ \\
    \parbox[t]{8cm}{\texttt{Aortic valve disorders}} & $0.8828\ (0.7213,\ 1.080)$ & $0.2266$ \\
    \parbox[t]{8cm}{\texttt{Lymphocytes\_max}} & $0.9934\ (0.9835,\ 1.0034)$ & $0.1944$ \\
    \parbox[t]{8cm}{\texttt{Other chronic pulmonary heart diseases}} & $1.070\ (0.9202,\ 1.244)$ & $0.3791$ \\
    \parbox[t]{8cm}{\texttt{Atrial fibrillation}} & $0.9931\ (0.8683,\ 1.136)$ & $0.9194$ \\
    \parbox[t]{8cm}{\texttt{Creatinine\_mean}} & $1.013\ (0.9562,\ 1.074)$ & $0.6561$ \\
    \parbox[t]{8cm}{\texttt{Urinary tract infection, site not specified}} & $0.8642\ (0.7574,\ 0.9862)$ & $0.0303$ \\
    \bottomrule
  \end{tabular}
  }
  }
\end{table}

\begin{table}[htbp]
\floatconts
  {tab:cox_sum_rsnt}
  {\caption{Hazard ratios, 95\% confidence intervals (CI), and p-values for ResNet-augmented Cox model.}}
  {\centering
  \resizebox{\columnwidth}{!}{
  % \small
  \begin{tabular}{l c c}
    \toprule
    \textbf{Features} & \textbf{Hazard Ratio (95\% CI)} & \textbf{p-value} \\
    \midrule
    \parbox[t]{8cm}{\texttt{Coronary atherosclerosis \\ of native coronary artery}} & $0.8884\ (0.7719,\ 1.0225)$ & $0.0989$ \\
    \parbox[t]{8cm}{\texttt{Renal dialysis status}} & $2.0339\ (1.6719,\ 2.4744)$ & $1.26 \times 10^{-12}$ \\
    \parbox[t]{8cm}{\texttt{Creatinine\_last}} & $0.9853\ (0.9410,\ 1.0318)$ & $0.5294$ \\
    \parbox[t]{8cm}{\texttt{MCHC\_min}} & $1.0140\ (0.9577,\ 1.0737)$ & $0.6331$ \\
    \parbox[t]{8cm}{\texttt{Anemia, unspecified}} & $0.9505\ (0.8341,\ 1.0831)$ & $0.4461$ \\
    \parbox[t]{8cm}{\texttt{Urea\_Nitrogen\_max}} & $0.9996\ (0.9961,\ 1.0031)$ & $0.8333$ \\
    \parbox[t]{8cm}{\texttt{marital\_status\_SINGLE}} & $1.1052\ (0.9317,\ 1.3110)$ & $0.2509$ \\
    \parbox[t]{8cm}{\texttt{Chloride\_max}} & $1.0004\ (0.9833,\ 1.0178)$ & $0.9630$ \\
    \parbox[t]{8cm}{\texttt{race\_WHITE}} & $0.9620\ (0.8415,\ 1.0997)$ & $0.5700$ \\
    \parbox[t]{8cm}{\texttt{Potassium\_min}} & $0.8582\ (0.6767,\ 1.0884)$ & $0.2072$ \\
    \parbox[t]{8cm}{\texttt{Unspecified essential hypertension}} & $0.6198\ (0.5425,\ 0.7081)$ & $1.90 \times 10^{-12}$ \\
    \parbox[t]{8cm}{\texttt{Urea\_Nitrogen\_mean}} & $1.0251\ (1.0064,\ 1.0441)$ & $0.0084$ \\
    \parbox[t]{8cm}{\texttt{Urea\_Nitrogen\_min}} & $1.0064\ (0.9969,\ 1.0159)$ & $0.1861$ \\
    \parbox[t]{8cm}{\texttt{Platelet\_Count\_last}} & $1.0005\ (0.9998,\ 1.0013)$ & $0.1419$ \\
    \parbox[t]{8cm}{\texttt{Urea\_Nitrogen\_median}} & $0.9929\ (0.9795,\ 1.0064)$ & $0.3003$ \\
    \parbox[t]{8cm}{\texttt{Esophageal reflux}} & $0.9256\ (0.8159,\ 1.0500)$ & $0.2295$ \\
    \parbox[t]{8cm}{\texttt{Other and unspecified hyperlipidemia}} & $1.0613\ (0.9205,\ 1.2236)$ & $0.4128$ \\
    \parbox[t]{8cm}{\texttt{Potassium\_last}} & $1.0992\ (0.9676,\ 1.2485)$ & $0.1459$ \\
    \parbox[t]{8cm}{\texttt{marital\_status\_MARRIED}} & $1.0400\ (0.8911,\ 1.2137)$ & $0.6188$ \\
    \parbox[t]{8cm}{\texttt{MCHC\_max}} & $0.9063\ (0.8517,\ 0.9643)$ & $0.0019$ \\
    \parbox[t]{8cm}{\texttt{Creatinine\_Urine\_max}} & $0.9996\ (0.9987,\ 1.0006)$ & $0.4338$ \\
    \parbox[t]{8cm}{\texttt{Monocytes\_min}} & $0.9704\ (0.9268,\ 1.0161)$ & $0.2010$ \\
    \parbox[t]{8cm}{\texttt{Potassium\_max}} & $0.9714\ (0.8131,\ 1.1604)$ & $0.7488$ \\
    \parbox[t]{8cm}{\texttt{Chloride\_first}} & $1.0020\ (0.9875,\ 1.0168)$ & $0.7880$ \\
    \parbox[t]{8cm}{\texttt{Lymphocytes\_max}} & $0.9883\ (0.9783,\ 0.9983)$ & $0.0226$ \\
    \parbox[t]{8cm}{\texttt{Hypertrophy of prostate without \\ urinary obstruction}} & $1.0246\ (0.8614,\ 1.2187)$ & $0.7836$ \\
    \parbox[t]{8cm}{\texttt{Diabetes mellitus type II, not uncontrolled}} & $0.8598\ (0.7569,\ 0.9767)$ & $0.0202$ \\
    \parbox[t]{8cm}{\texttt{Peripheral vascular disease, unspecified}} & $0.9851\ (0.8448,\ 1.1487)$ & $0.8478$ \\
    \parbox[t]{8cm}{\texttt{Percutaneous transluminal coronary \\ angioplasty status}} & $1.0821\ (0.9169,\ 1.2770)$ & $0.3507$ \\
    \parbox[t]{8cm}{\texttt{Aortic valve disorders}} & $0.8724\ (0.7054,\ 1.0789)$ & $0.2080$ \\
    \parbox[t]{8cm}{\texttt{Physical restraints status}} & $0.7141\ (0.5096,\ 1.0008)$ & $0.0505$ \\
    \parbox[t]{8cm}{\texttt{Creatinine\_mean}} & $1.0826\ (1.0067,\ 1.1642)$ & $0.0323$ \\
    \parbox[t]{8cm}{\texttt{Platelet\_Count\_max}} & $0.9994\ (0.9989,\ 1.0000)$ & $0.0290$ \\
    \parbox[t]{8cm}{\texttt{Neutrophils\_last}} & $1.0063\ (1.0000,\ 1.0126)$ & $0.0494$ \\
    \parbox[t]{8cm}{\texttt{Neutrophils\_max}} & $0.9921\ (0.9816,\ 1.0028)$ & $0.1476$ \\
    \parbox[t]{8cm}{\texttt{Pure hypercholesterolemia}} & $0.9737\ (0.8515,\ 1.1134)$ & $0.6971$ \\
    \parbox[t]{8cm}{\texttt{MCV\_min}} & $1.0139\ (1.0036,\ 1.0243)$ & $0.0078$ \\
    \parbox[t]{8cm}{\texttt{Hypovolemia}} & $0.8641\ (0.7234,\ 1.0321)$ & $0.1071$ \\
    \parbox[t]{8cm}{\texttt{Neutrophils\_min}} & $0.9857\ (0.9765,\ 0.9949)$ & $0.0024$ \\
    \parbox[t]{8cm}{\texttt{Gout, unspecified}} & $0.9369\ (0.8115,\ 1.0818)$ & $0.3747$ \\
    \bottomrule
  \end{tabular}
  }
  }
\end{table}

\begin{table}[htbp]
\floatconts
  {tab:cox_sum_lr}
  {\caption{Hazard ratios, 95\% confidence intervals (CI), and p-values for LR-augmented Cox model.}}
  {\centering
  \resizebox{\columnwidth}{!}{
  \small
  \begin{tabular}{l c c}
    \toprule
    \textbf{Features} & \textbf{Hazard Ratio (95\% CI)} & \textbf{p-value} \\
    \midrule
    \texttt{Renal dialysis status} & $2.0084\ (1.6313,\ 2.4728)$ & $4.99 \times 10^{-11}$ \\
    \texttt{Creatinine\_last} & $0.9998\ (0.9538,\ 1.0480)$ & $0.9935$ \\
    \texttt{Urea\_Nitrogen\_mean} & $1.0154\ (1.0069,\ 1.0240)$ & $0.00036$ \\
    \texttt{MCH\_min} & $1.0939\ (1.0341,\ 1.1572)$ & $0.0018$ \\
    \texttt{MCHC\_min} & $1.0058\ (0.9336,\ 1.0837)$ & $0.8784$ \\
    \texttt{Protein\_mean} & $1.0013\ (1.0006,\ 1.0020)$ & $0.00053$ \\
    \texttt{Cholesterol\_Ratio\_ (Total/HDL)\_first} & $1.0189\ (0.9661,\ 1.0746)$ & $0.4899$ \\
    \texttt{Nucleated\_Red\_Cells\_last} & $1.0078\ (1.0001,\ 1.0155)$ & $0.0460$ \\
    \texttt{Basophils\_mean} & $1.0327\ (0.7883,\ 1.3530)$ & $0.8153$ \\
    \texttt{Chloride\_max} & $0.9987\ (0.9789,\ 1.0190)$ & $0.9009$ \\
    \texttt{Unspecified essential hypertension} & $0.5907\ (0.5172,\ 0.6747)$ & $8.35 \times 10^{-15}$ \\
    \texttt{Protein\_Total\_last} & $1.1584\ (0.9418,\ 1.4249)$ & $0.1640$ \\
    \texttt{Chloride\_last} & $1.0061\ (0.9912,\ 1.0211)$ & $0.4251$ \\
    \texttt{Urea\_Nitrogen\_last} & $0.9992\ (0.9959,\ 1.0026)$ & $0.6607$ \\
    \texttt{Protein\_Total\_max} & $0.8192\ (0.6546,\ 1.0252)$ & $0.0814$ \\
    \texttt{Cholesterol\_HDL\_last} & $0.9975\ (0.9923,\ 1.0027)$ & $0.3405$ \\
    \texttt{Cholesterol\_Total\_first} & $0.9984\ (0.9966,\ 1.0003)$ & $0.0932$ \\
    \texttt{Uric\_Acid\_median} & $1.0217\ (0.9822,\ 1.0628)$ & $0.2853$ \\
    \texttt{Urea\_Nitrogen\_max} & $1.0005\ (0.9973,\ 1.0037)$ & $0.7593$ \\
    \texttt{MCH\_mean} & $0.8499\ (0.7670,\ 0.9417)$ & $0.0019$ \\
    \texttt{Other and unspecified hyperlipidemia} & $0.9549\ (0.8326,\ 1.0951)$ & $0.5089$ \\
    \texttt{marital\_status\_SINGLE} & $1.1004\ (0.9573,\ 1.2650)$ & $0.1784$ \\
    \texttt{MCV\_mean} & $1.0363\ (1.0024,\ 1.0715)$ & $0.0357$ \\
    \texttt{Sodium\_mean} & $0.9867\ (0.9545,\ 1.0199)$ & $0.4275$ \\
    \texttt{Anemia, unspecified} & $0.9385\ (0.8235,\ 1.0694)$ & $0.3406$ \\
    \texttt{Atrial fibrillation} & $0.9819\ (0.8596,\ 1.1214)$ & $0.7872$ \\
    \texttt{Severe sepsis} & $0.8669\ (0.7257,\ 1.0357)$ & $0.1156$ \\
    \texttt{Creatinine\_median} & $1.0763\ (1.0056,\ 1.1519)$ & $0.0339$ \\
    \texttt{Creatinine\_first} & $0.9238\ (0.8646,\ 0.9871)$ & $0.0190$ \\
    \texttt{Myelocytes\_mean} & $1.0185\ (0.9054,\ 1.1457)$ & $0.7602$ \\
    \texttt{Creatinine\_Urine\_last} & $1.0002\ (0.9990,\ 1.0014)$ & $0.7027$ \\
    \texttt{Neutrophils\_min} & $0.9975\ (0.9927,\ 1.0024)$ & $0.3178$ \\
    \parbox[t]{8cm}{\texttt{Diabetes with renal manifestations,\\ type II or unspecified type}} & $0.8890\ (0.7646,\ 1.0336)$ & $0.1261$ \\
    \texttt{Protein\_Total\_first} & $0.9964\ (0.8087,\ 1.2276)$ & $0.9727$ \\
    \texttt{Globulin\_last} & $1.1550\ (0.9877,\ 1.3507)$ & $0.0710$ \\
    \texttt{Urea\_Nitrogen\_min} & $1.0090\ (0.9994,\ 1.0186)$ & $0.0660$ \\
    \texttt{Cholesterol\_LDL\_Calculated\_median} & $1.0027\ (1.0000,\ 1.0053)$ & $0.0458$ \\
    \texttt{Monocytes\_min} & $0.9893\ (0.9453,\ 1.0353)$ & $0.6425$ \\
    \texttt{Neutrophils\_max} & $0.9973\ (0.9877,\ 1.0070)$ & $0.5862$ \\
    \texttt{Protein\_Total\_min} & $0.9026\ (0.7470,\ 1.0907)$ & $0.2887$ \\
    \bottomrule
  \end{tabular}
  }
  }
\end{table}

\begin{table}[htbp]
\floatconts
  {tab:cox_sum_dt}
  {\caption{Hazard ratios, 95\% confidence intervals (CI), and p-values for DT-augmented Cox model.}}
  {\centering
  \resizebox{\columnwidth}{!}{
  \small
  \begin{tabular}{l c c}
    \toprule
    \textbf{Features} & \textbf{Hazard Ratio (95\% CI)} & \textbf{p-value} \\
    \midrule
    \texttt{Creatinine\_max} & $0.9517\ (0.9167,\ 0.9880)$ & $0.0095$ \\
    \texttt{Renal dialysis status} & $1.8287\ (1.5099,\ 2.2148)$ & $6.58 \times 10^{-10}$ \\
    \texttt{Creatinine\_mean} & $1.2372\ (1.0643,\ 1.4383)$ & $0.0056$ \\
    \texttt{Creatinine\_last} & $1.0405\ (0.9948,\ 1.0884)$ & $0.0834$ \\
    \texttt{Creatinine\_median} & $0.9259\ (0.8307,\ 1.0320)$ & $0.1641$ \\
    \texttt{RBC\_max} & $1.0001\ (0.9996,\ 1.0007)$ & $0.6106$ \\
    \texttt{Urea\_Nitrogen\_max} & $0.9988\ (0.9956,\ 1.0020)$ & $0.4569$ \\
    \texttt{Protein\_first} & $1.0010\ (1.0005,\ 1.0014)$ & $3.77 \times 10^{-5}$ \\
    \texttt{Monocytes\_first} & $1.0064\ (0.9816,\ 1.0319)$ & $0.6162$ \\
    \texttt{Platelet\_Count\_first} & $0.9996\ (0.9989,\ 1.0003)$ & $0.2354$ \\
    \texttt{Diabetes with other specified manifestations} & $0.9198\ (0.7790,\ 1.0861)$ & $0.3241$ \\
    \texttt{Uric\_Acid\_max} & $0.9793\ (0.9481,\ 1.0116)$ & $0.2065$ \\
    \texttt{Acute kidney failure, unspecified} & $1.0816\ (0.9249,\ 1.2649)$ & $0.3257$ \\
    \texttt{Urea\_Nitrogen\_mean} & $1.0217\ (1.0151,\ 1.0284)$ & $1.26 \times 10^{-10}$ \\
    \texttt{MCHC\_mean} & $0.9559\ (0.9003,\ 1.0149)$ & $0.1399$ \\
    \texttt{Pure hypercholesterolemia} & $0.8793\ (0.7736,\ 0.9995)$ & $0.0491$ \\
    \texttt{Absolute\_Lymphocyte\_Count\_last} & $0.8964\ (0.7646,\ 1.0510)$ & $0.1780$ \\
    \texttt{Alanine\_Aminotransferase\ (ALT)\_median} & $0.9991\ (0.9980,\ 1.0002)$ & $0.0982$ \\
    \texttt{Other musculoskeletal symptoms} & $0.9351\ (0.7290,\ 1.1995)$ & $0.5974$ \\
    \texttt{Cholesterol\_Ratio\_(Total/HDL)\_last} & $1.0376\ (0.9962,\ 1.0807)$ & $0.0755$ \\
    \texttt{Glucose\_mean} & $0.9964\ (0.9945,\ 0.9984)$ & $3.11 \times 10^{-4}$ \\
    \texttt{Platelet\_Count\_last} & $1.0001\ (0.9995,\ 1.0008)$ & $0.6477$ \\
    \texttt{Reticulocyte\_Count\_Manual\_min} & $1.0371\ (0.9430,\ 1.1405)$ & $0.4533$ \\
    \bottomrule
  \end{tabular}
  }
  }
\end{table}

\pagebreak

\begin{table}[htbp]
\floatconts
  {tab:cox_sum_rf}
  {\caption{Hazard ratios, 95\% confidence intervals (CI), and p-values for RF-augmented Cox model.}}
  {\centering
  \resizebox{\columnwidth}{!}{
  \small
  \begin{tabular}{l c c}
    \toprule
    \textbf{Features} & \textbf{Hazard Ratio (95\% CI)} & \textbf{p-value} \\
    \midrule
    \texttt{Urea\_Nitrogen\_max}           & $0.9990\ (0.9956,\ 1.0024)$  & $0.5460$ \\
    \texttt{Creatinine\_max}               & $0.9083\ (0.8654,\ 0.9533)$  & $9.81 \times 10^{-5}$ \\
    \texttt{Creatinine\_last}              & $1.0238\ (0.9762,\ 1.0738)$  & $0.3323$ \\
    \texttt{Creatinine\_mean}              & $1.5034\ (1.2271,\ 1.8420)$  & $8.31 \times 10^{-5}$ \\
    \texttt{Renal dialysis status}         & $2.1128\ (1.7321,\ 2.5771)$  & $1.58 \times 10^{-13}$ \\
    \texttt{Creatinine\_median}            & $0.8243\ (0.7081,\ 0.9596)$  & $0.0127$ \\
    \texttt{Potassium\_max}                & $0.9503\ (0.8078,\ 1.1178)$  & $0.5382$ \\
    \texttt{Urea\_Nitrogen\_mean}          & $1.0280\ (1.0106,\ 1.0456)$  & $0.0015$ \\
    \texttt{Protein/Creatinine\_Ratio\_last} & $1.0250\ (0.9944,\ 1.0567)$  & $0.1108$ \\
    \texttt{Eosinophils\_max}              & $0.9956\ (0.9831,\ 1.0082)$  & $0.4928$ \\
    \texttt{Creatinine\_Urine\_min}        & $0.9995\ (0.9976,\ 1.0014)$  & $0.5852$ \\
    \texttt{Total Protein\_Urine\_mean}    & $0.9973\ (0.9923,\ 1.0022)$  & $0.2794$ \\
    \texttt{MCHC\_min}                     & $1.0139\ (0.9657,\ 1.0645)$  & $0.5788$ \\
    \texttt{Protein/Creatinine\_Ratio\_median} & $1.1237\ (0.9762,\ 1.2936)$  & $0.1043$ \\
    \texttt{Total Protein\_Urine\_max}     & $0.9997\ (0.9985,\ 1.0009)$  & $0.6200$ \\
    \texttt{Hemoglobin\_min\_y}            & $0.9747\ (0.9311,\ 1.0204)$  & $0.2731$ \\
    \texttt{Protein/Creatinine\_Ratio\_mean} & $0.9218\ (0.7635,\ 1.1128)$  & $0.3965$ \\
    \texttt{Uric\_Acid\_mean}              & $0.9381\ (0.7166,\ 1.2280)$  & $0.6419$ \\
    \texttt{Total Protein\_Urine\_median}  & $1.0014\ (0.9983,\ 1.0045)$  & $0.3690$ \\
    \texttt{Acute kidney failure, unspecified} & $1.0813\ (0.9225,\ 1.2673)$  & $0.3348$ \\
    \texttt{Protein/Creatinine\_Ratio\_max} & $0.9969\ (0.9587,\ 1.0367)$  & $0.8782$ \\
    \texttt{Urea\_Nitrogen\_Urine\_min}    & $1.0001\ (0.9995,\ 1.0007)$  & $0.7437$ \\
    \texttt{Protein\_mean}                 & $1.0020\ (1.0008,\ 1.0032)$  & $0.0014$ \\
    \texttt{Urea\_Nitrogen\_median}        & $0.9950\ (0.9816,\ 1.0086)$  & $0.4723$ \\
    \texttt{Asparate\_Aminotransferase\ (AST)\_min} & $0.9996\ (0.9982,\ 1.0010)$ & $0.5581$ \\
    \texttt{Total Protein\_Urine\_last}    & $1.0005\ (0.9995,\ 1.0016)$  & $0.3179$ \\
    \texttt{Protein\_max}                  & $0.9994\ (0.9988,\ 1.0000)$  & $0.0632$ \\
    \texttt{Uric\_Acid\_max}               & $0.9566\ (0.8956,\ 1.0218)$  & $0.1875$ \\
    \texttt{Uric\_Acid\_median}            & $1.1060\ (0.8992,\ 1.3604)$  & $0.3402$ \\
    \texttt{Protein/Creatinine\_Ratio\_first} & $1.0372\ (0.9854,\ 1.0917)$  & $0.1620$ \\
    \texttt{Total Protein\_Urine\_first}   & $1.0009\ (0.9996,\ 1.0021)$  & $0.1653$ \\
    \texttt{Total Protein\_Urine\_min}     & $1.0013\ (0.9992,\ 1.0034)$  & $0.2300$ \\
    \texttt{Chloride\_min}                 & $1.0189\ (1.0048,\ 1.0333)$  & $0.0085$ \\
    \texttt{Protein\_last}                 & $0.9996\ (0.9989,\ 1.0002)$  & $0.1891$ \\
    \texttt{Uric\_Acid\_last}              & $1.0389\ (0.9678,\ 1.1151)$  & $0.2912$ \\
    \texttt{Albumin\_Urine\_max}           & $0.9989\ (0.9969,\ 1.0010)$  & $0.3093$ \\
    \texttt{Neutrophils\_min}              & $0.9978\ (0.9930,\ 1.0025)$  & $0.3541$ \\
    \texttt{Protein/Creatinine\_Ratio\_min} & $0.9316\ (0.8552,\ 1.0147)$  & $0.1042$ \\
    \texttt{Urea\_Nitrogen\_Urine\_last}   & $0.9995\ (0.9989,\ 0.9999)$  & $0.0437$ \\
    \texttt{Albumin\_Urine\_last}          & $1.0005\ (0.9981,\ 1.0029)$  & $0.6867$ \\
    \bottomrule
  \end{tabular}
  }
  }
\end{table}

\pagebreak

% This is the first appendix.

% MOVE THIS SOMEWHERE ELSE!!

% \begin{figure}[htbp]
%     \floatconts
%     {fig:shap_beeswarm} % Main figure label
%     {\caption{SHAP beeswarm plots for the top 40 features from feature selection models, ranked by the highest SHAPma values.}}
%     {
%         \subfigure[XGboost SHAP beeswarm plot]{\label{fig:XGboost_beeswarm}\includegraphics[width=0.75\linewidth]{images/xgb_Beeswarm.png}}\\ % Stack images vertically
%         \subfigure[FCNN SHAP beeswarm plot]{\label{fig:FCNN_beeswarm}\includegraphics[width=\linewidth]{images/fcnn_Beeswarm.png}}\\ % Stack images vertically
%         \subfigure[ResNet SHAP beeswarm plot]{\label{fig:ResNet_beeswarm}\includegraphics[width=\linewidth]{images/rsnt_Beeswarm.png}}   % Stack images vertically
%     }
% \end{figure}

%% file: 9-appendix-F-mean-shap-val.tex
\section{Mean Absolute SHAP Values}\label{apd:shapma}

\begin{figure*}[htbp]
    \floatconts
    {fig:xgboost_shapma} % Label goes here
    {\caption{Top 40 mean absolute SHAP values from XGboost model.}}
    {\includegraphics[width=0.6\linewidth]{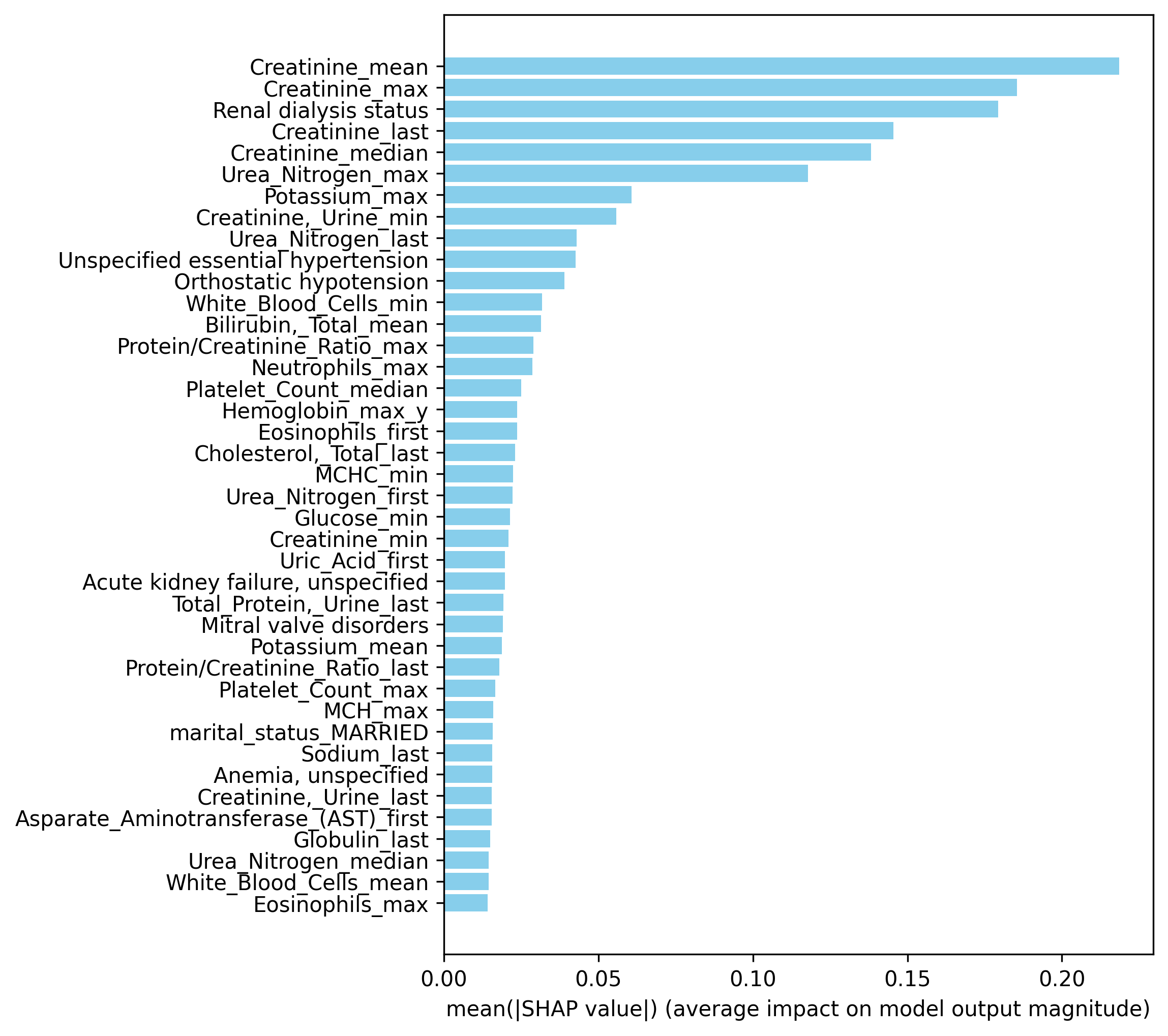}}
\end{figure*}

\begin{figure*}[htbp]
    \floatconts
    {fig:fcnn_shapma} % Label goes here
    {\caption{Top 40 mean absolute SHAP values from FCNN model.}}
    {\includegraphics[width=\linewidth]{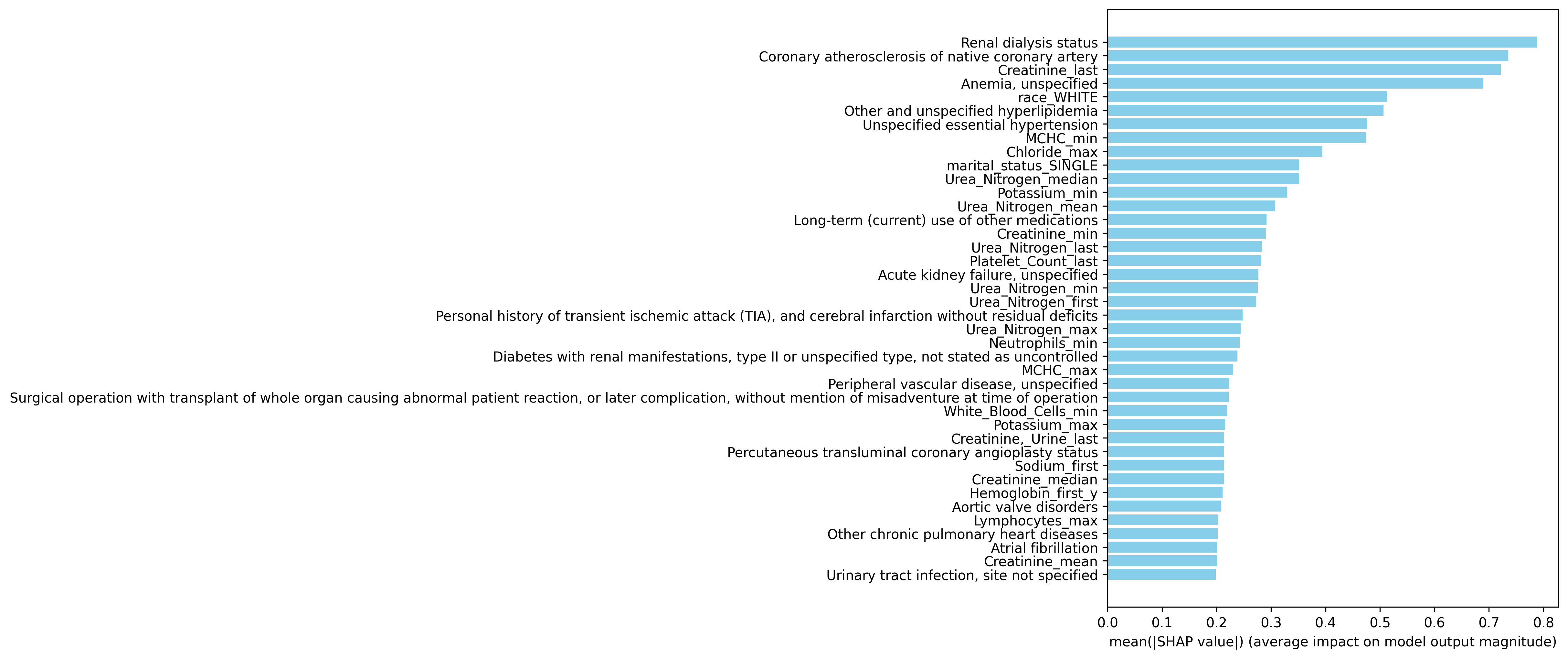}}
\end{figure*}

\begin{figure*}[htbp]
    \floatconts
    {fig:resnet_shapma} % Label goes here
    {\caption{Top 40 mean absolute SHAP values from ResNet model.}}
    {\includegraphics[width=\linewidth]{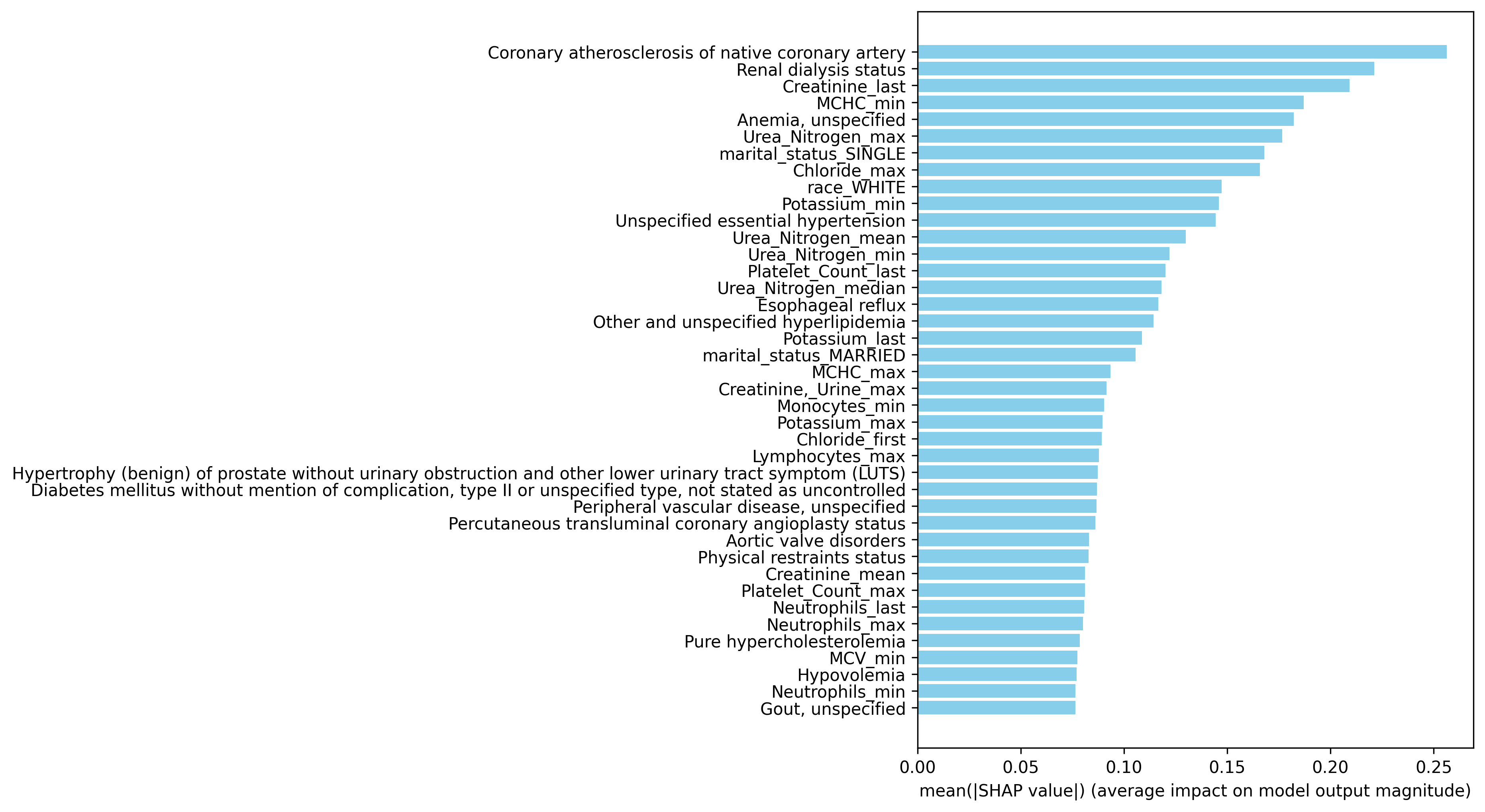}}
\end{figure*}

\begin{figure*}[htbp]
    \floatconts
    {fig:lr_shapma} % Label goes here
    {\caption{Top 40 mean absolute SHAP values from LR model.}}
    {\includegraphics[width=0.925\linewidth]{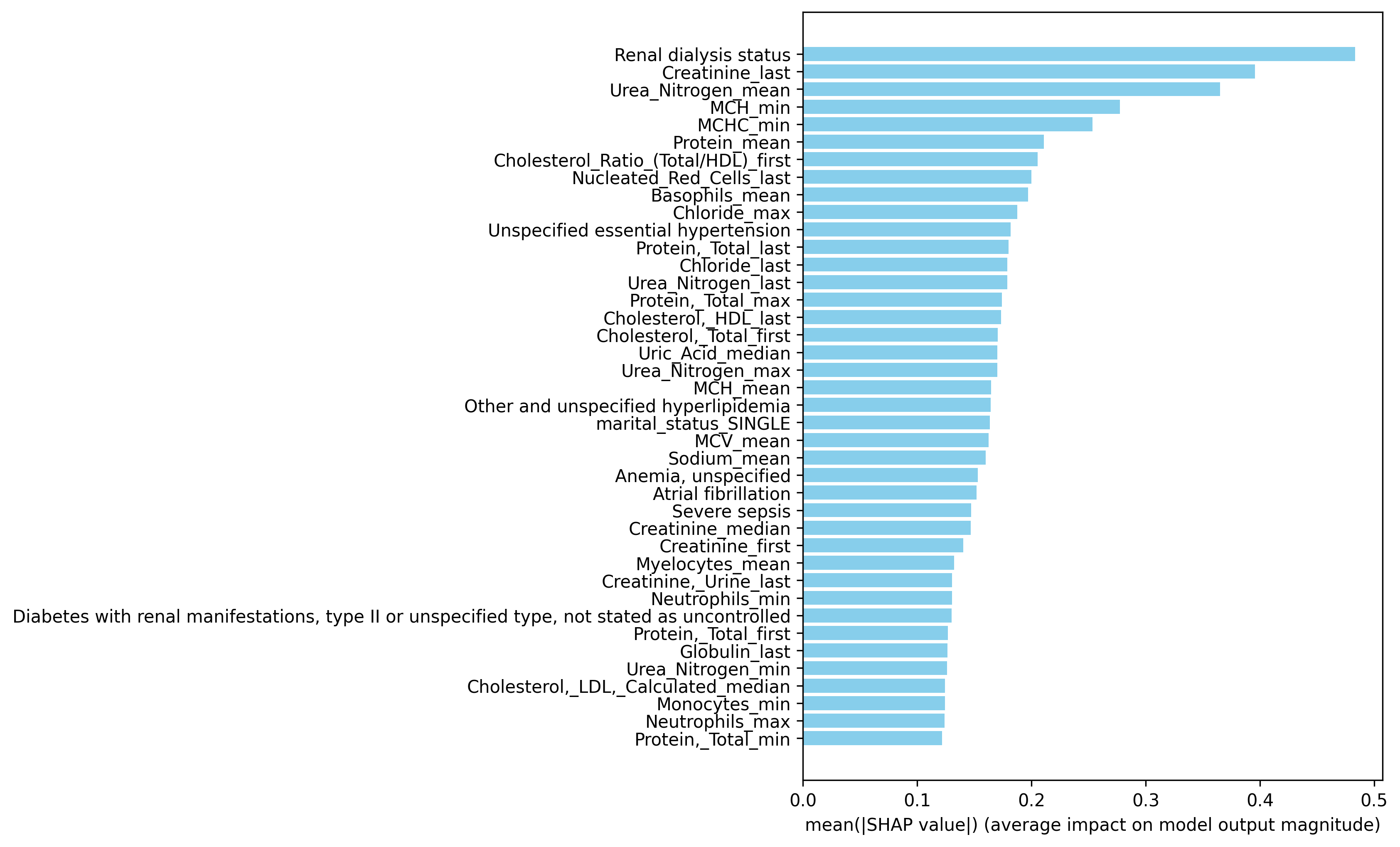}}
\end{figure*}

\begin{figure*}[htbp]
    \floatconts
    {fig:dt_shapma} % Label goes here
    {\caption{Top 40 mean absolute SHAP values from DT model.}}
    {\includegraphics[width=0.95\linewidth]{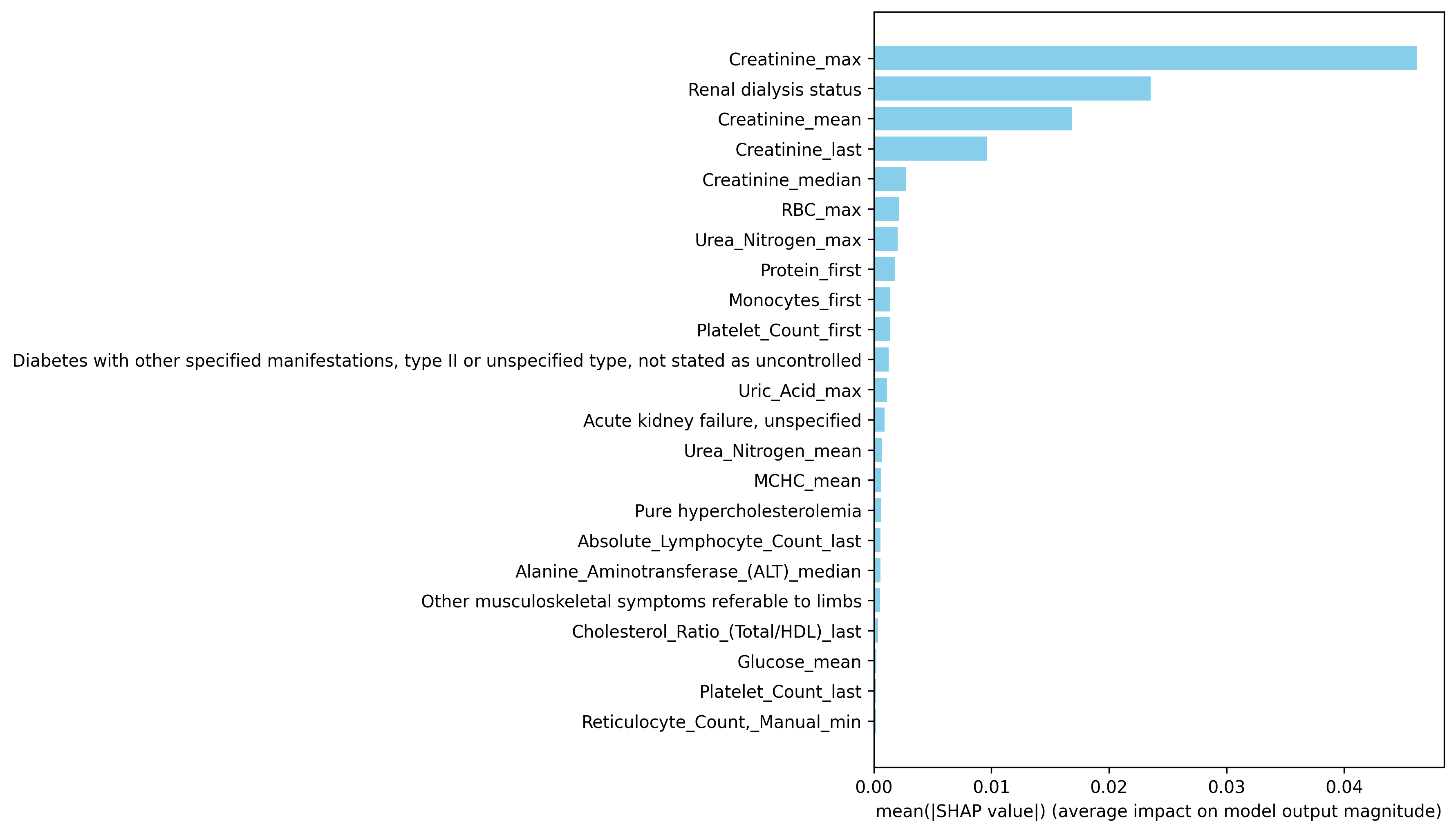}}
\end{figure*}

\begin{figure*}[htbp]
    \floatconts
    {fig:rf_shapma} % Label goes here
    {\caption{Top 40 mean absolute SHAP values from RF model.}}
    {\includegraphics[width=0.65\linewidth]{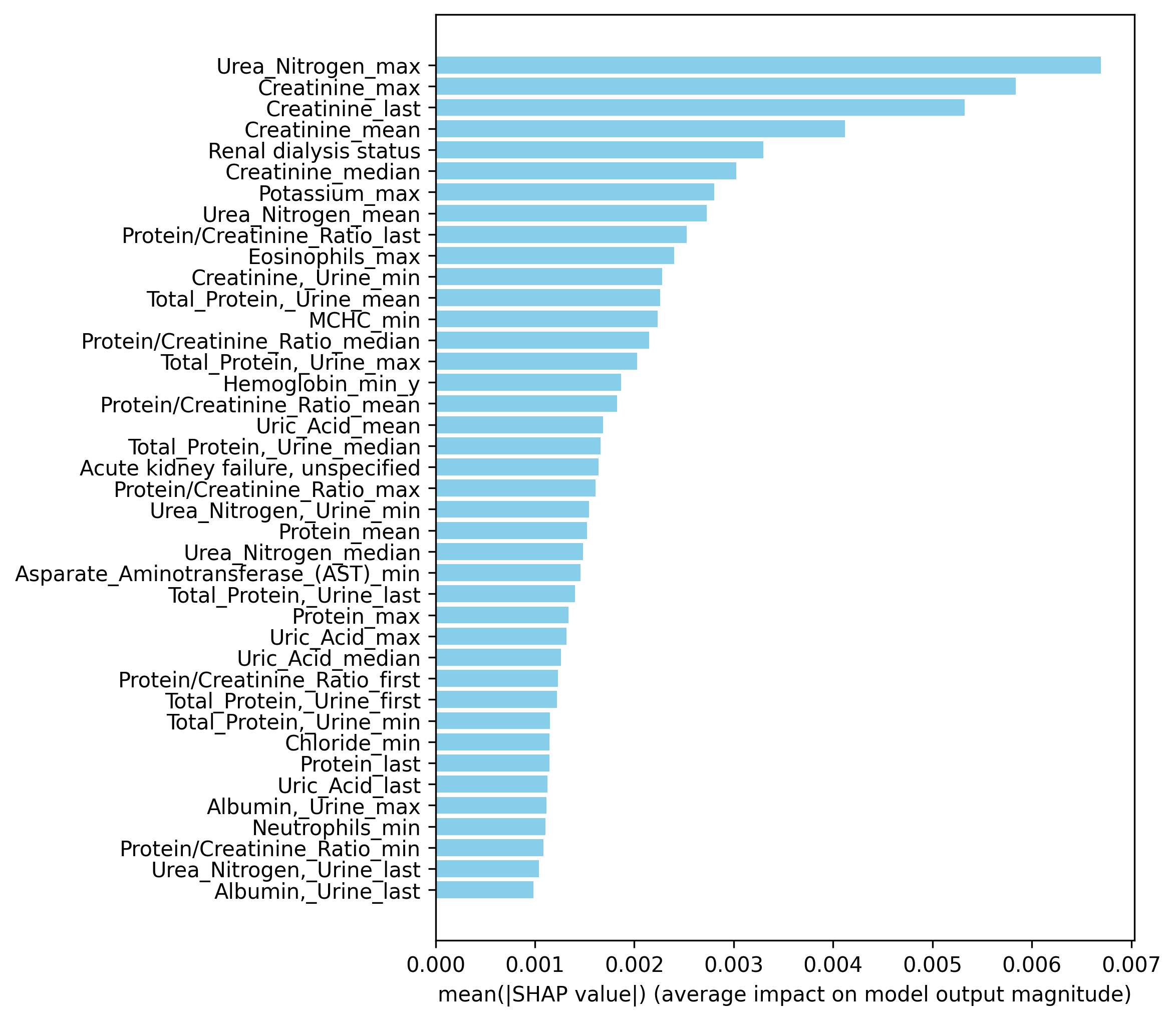}}
\end{figure*}

% \begin{figure}[htbp]
%     \centering
%     \includegraphics[width=\textwidth]{TripleVennFeatures.png}
%     \caption{Triple Venn Diagram: Top 40 Features Identified by Feature Selection Models}
%     \label{fig:triple_venn_features}
% \end{figure}

% \newpage

\clearpage

%% file: 9-appendix-G-brier.tex
\section{Brier Scores}
\label{apd:brier}

% \newpage
 
\begin{figure}[htbp]
    \floatconts
    {fig:baseline_brier} % Label goes here
    {\caption{Brier score plot for baseline Cox model.}}
    {\includegraphics[width=\linewidth]{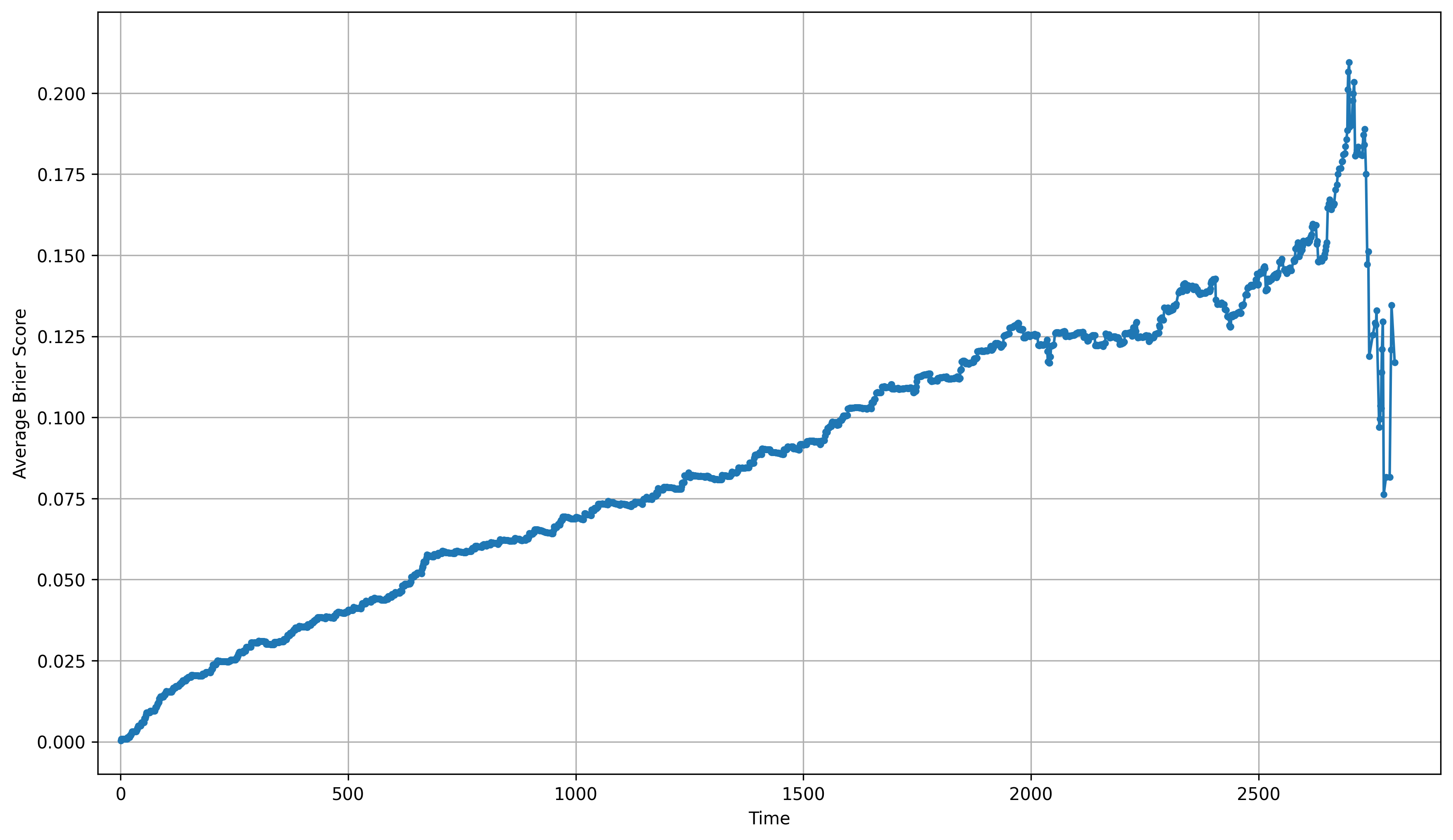}}
\end{figure}

\begin{figure}[htbp]
    \floatconts
    {fig:xgboost_brier} % Label goes here
    {\caption{Brier score plot for XGboost-augmented Cox model.}}
    {\includegraphics[width=\linewidth]{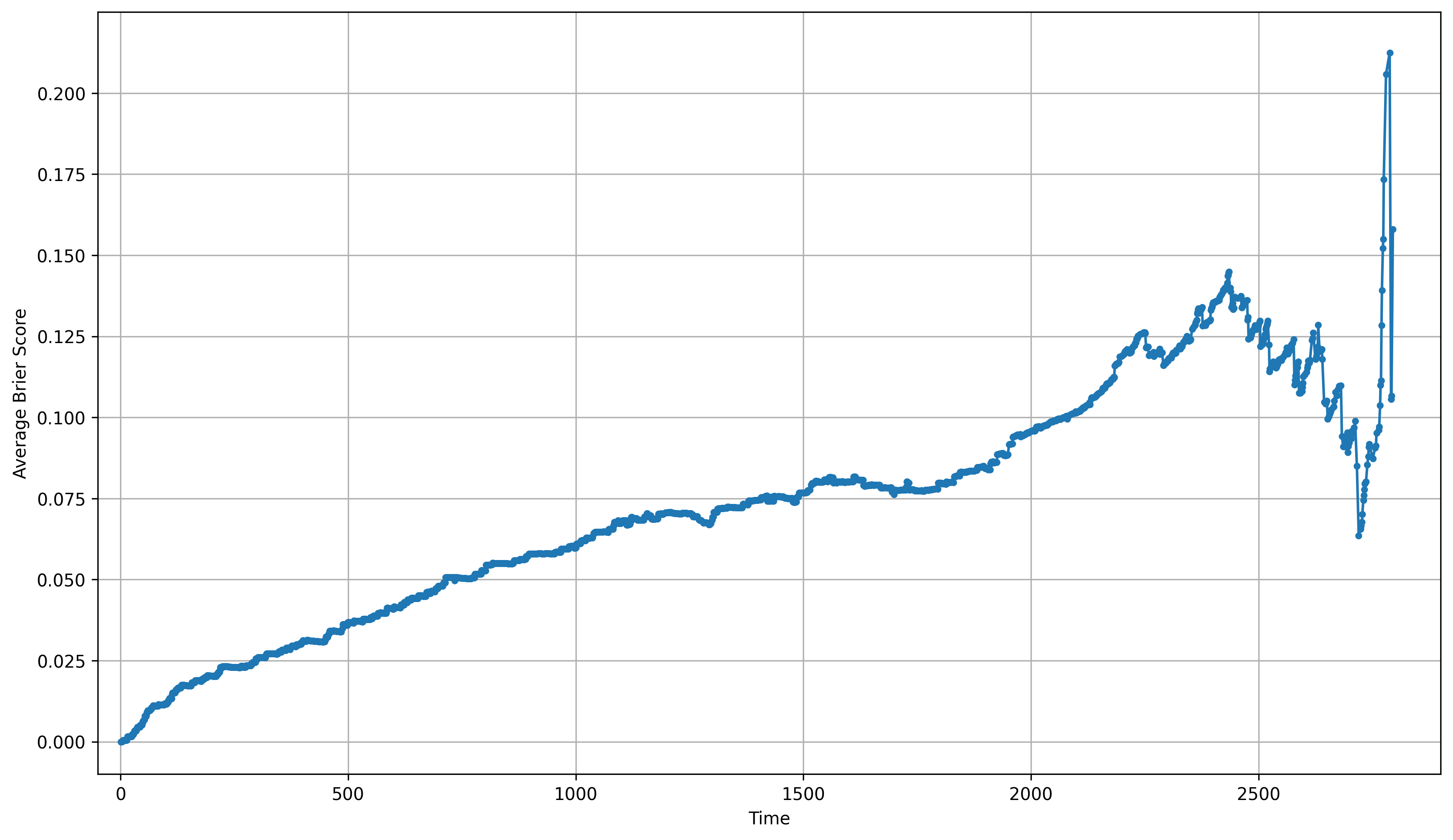}}
\end{figure}

\begin{figure}[htbp]
    \floatconts
    {fig:fcnn_brier} % Label goes here
    {\caption{Brier score plot for FCNN-augmented Cox model.}}
    {\includegraphics[width=\linewidth]{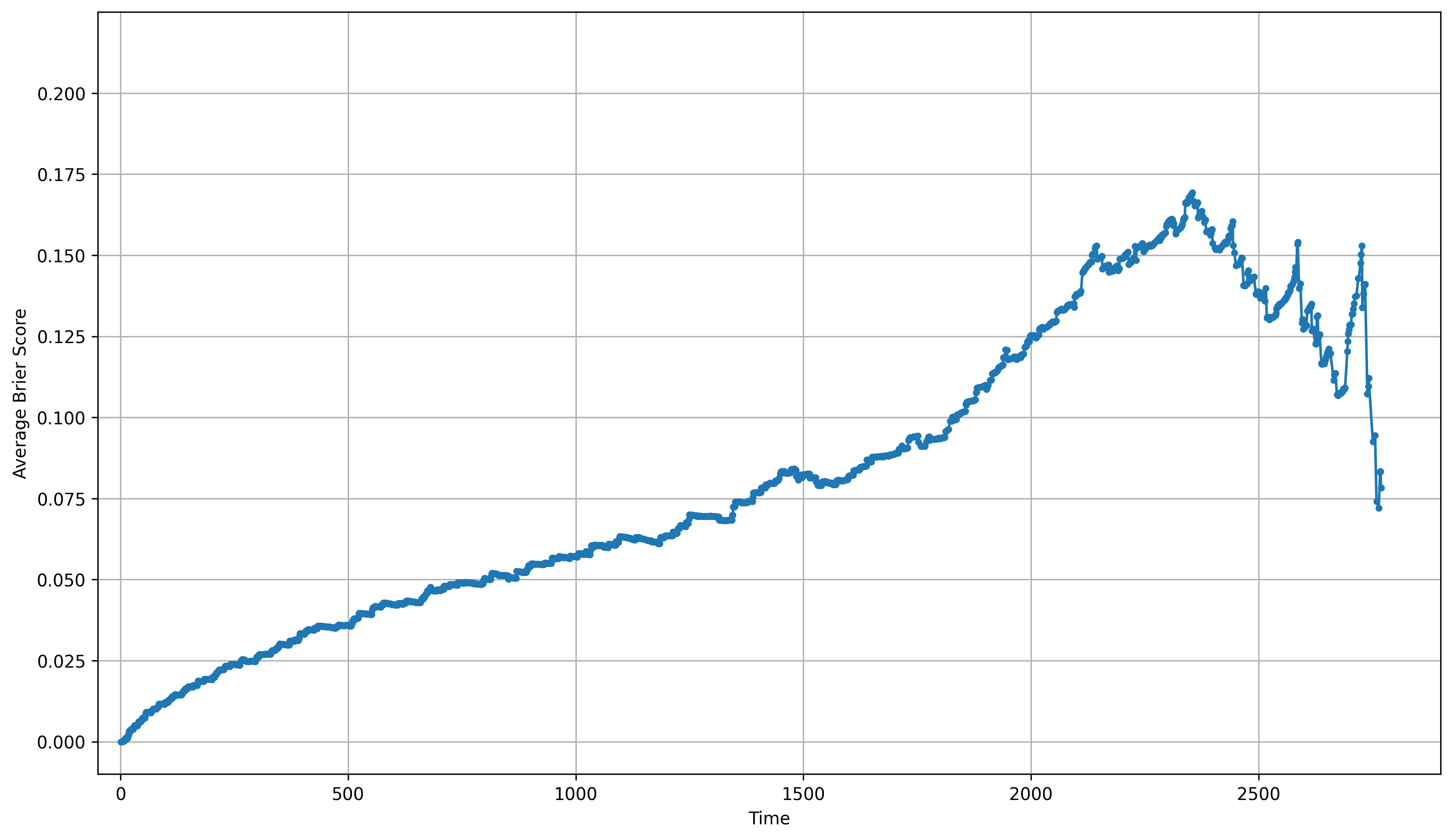}}
\end{figure}

\begin{figure}[htbp]
    \floatconts
    {fig:resnet_brier} % Label goes here
    {\caption{Brier score plot for ResNet-augmented Cox model.}}
    {\includegraphics[width=\linewidth]{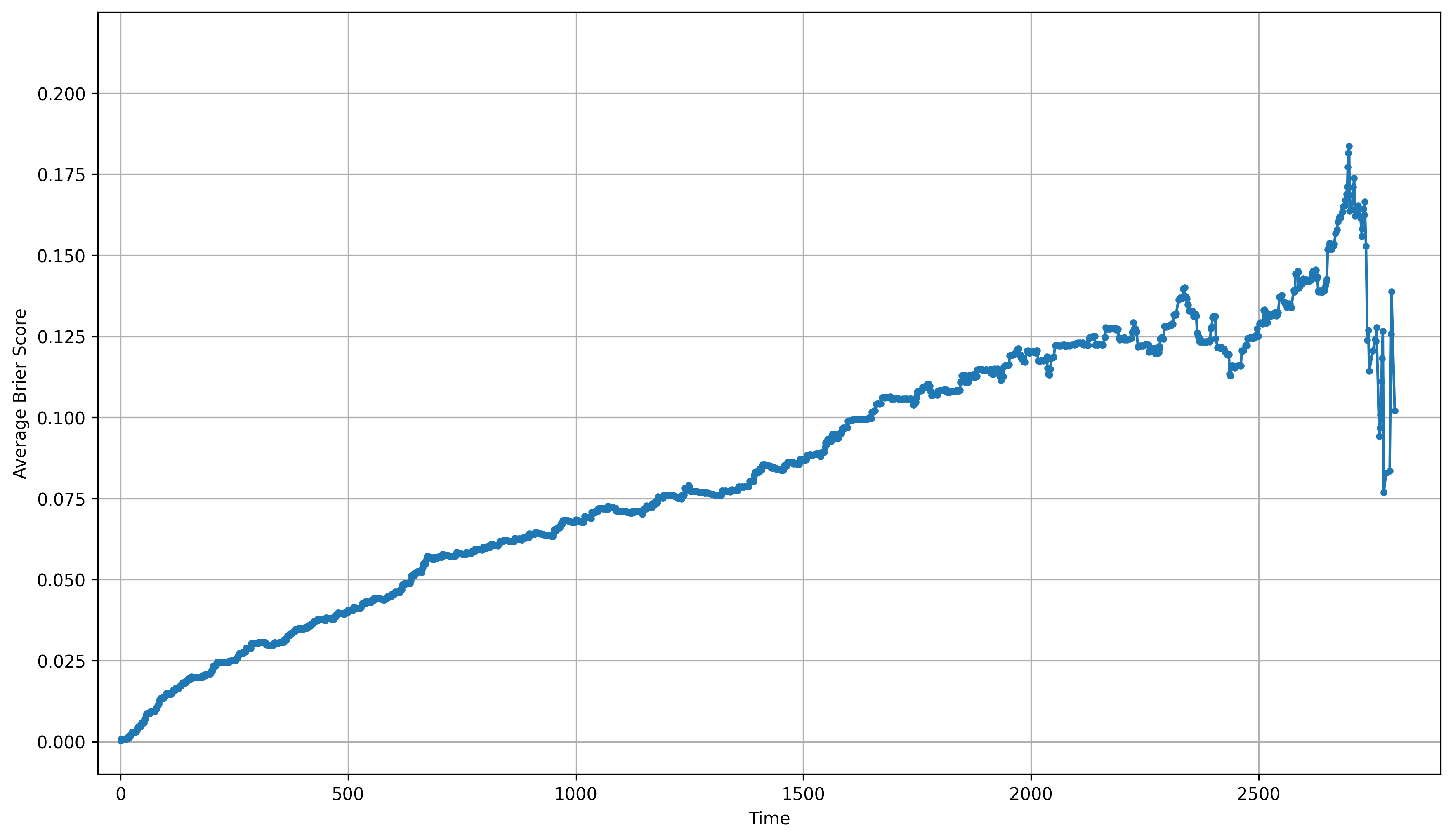}}
\end{figure}

\begin{figure}[htbp]
    \floatconts
    {fig:lr_brier} % Label goes here
    {\caption{Brier score plot for LR-augmented Cox model.}}
    {\includegraphics[width=\linewidth]{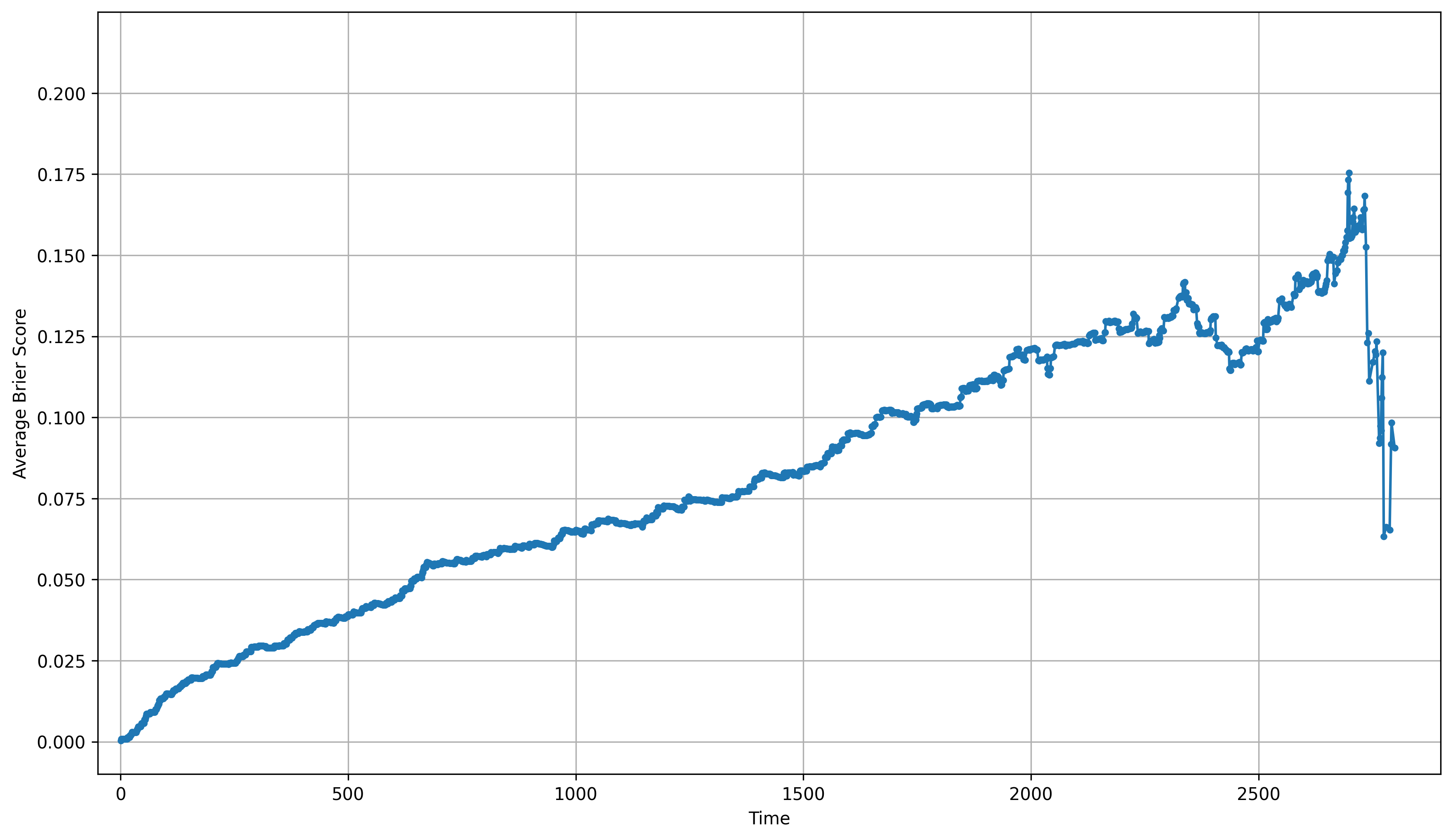}}
\end{figure}

\begin{figure}[htbp]
    \floatconts
    {fig:dt_brier} % Label goes here
    {\caption{Brier score plot for FCNN-augmented Cox model.}}
    {\includegraphics[width=\linewidth]{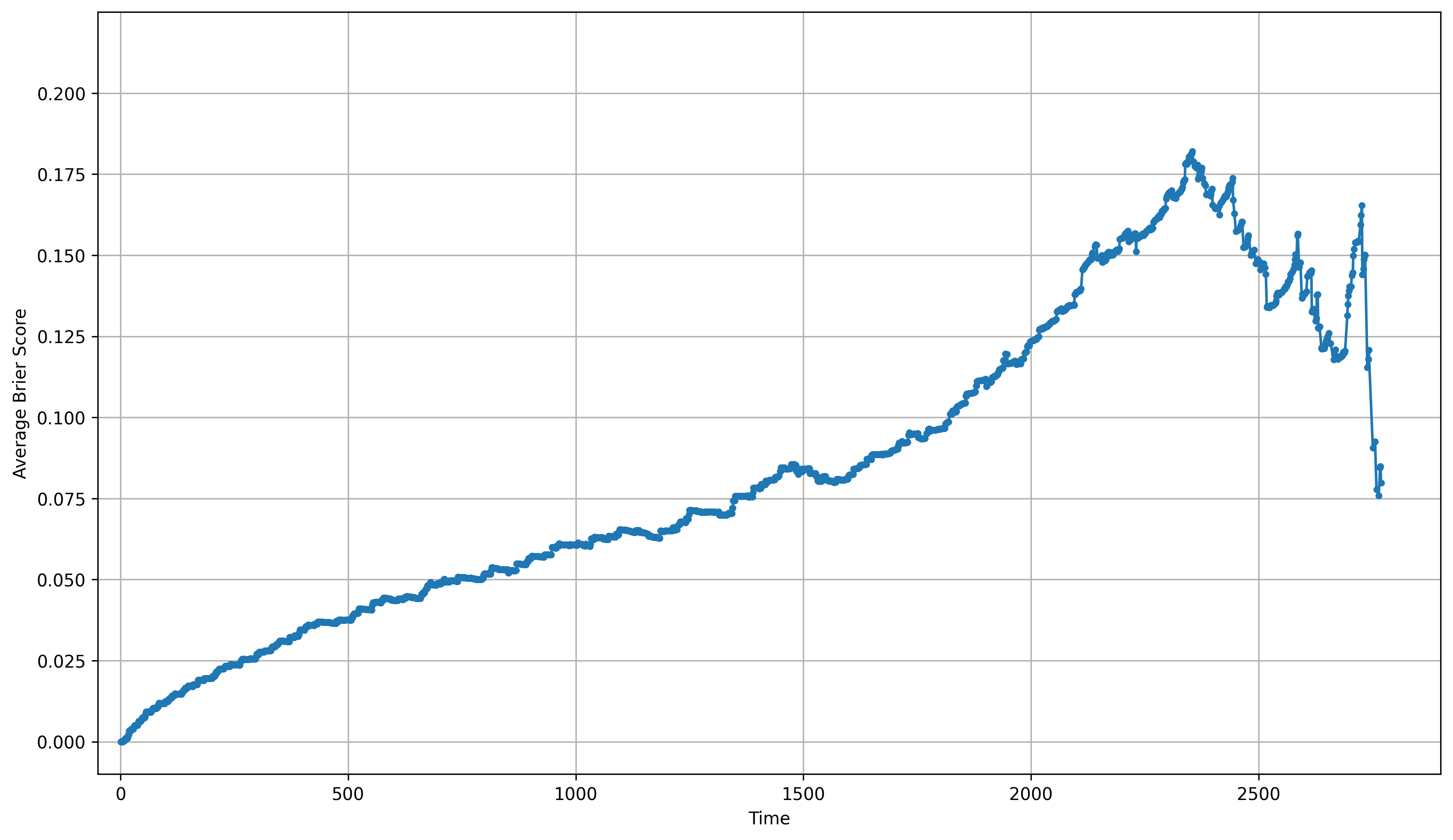}}
\end{figure}

\begin{figure}[htbp]
    \floatconts
    {fig:rf_brier} % Label goes here
    {\caption{Brier score plot for ResNet-augmented Cox model.}}
    {\includegraphics[width=\linewidth]{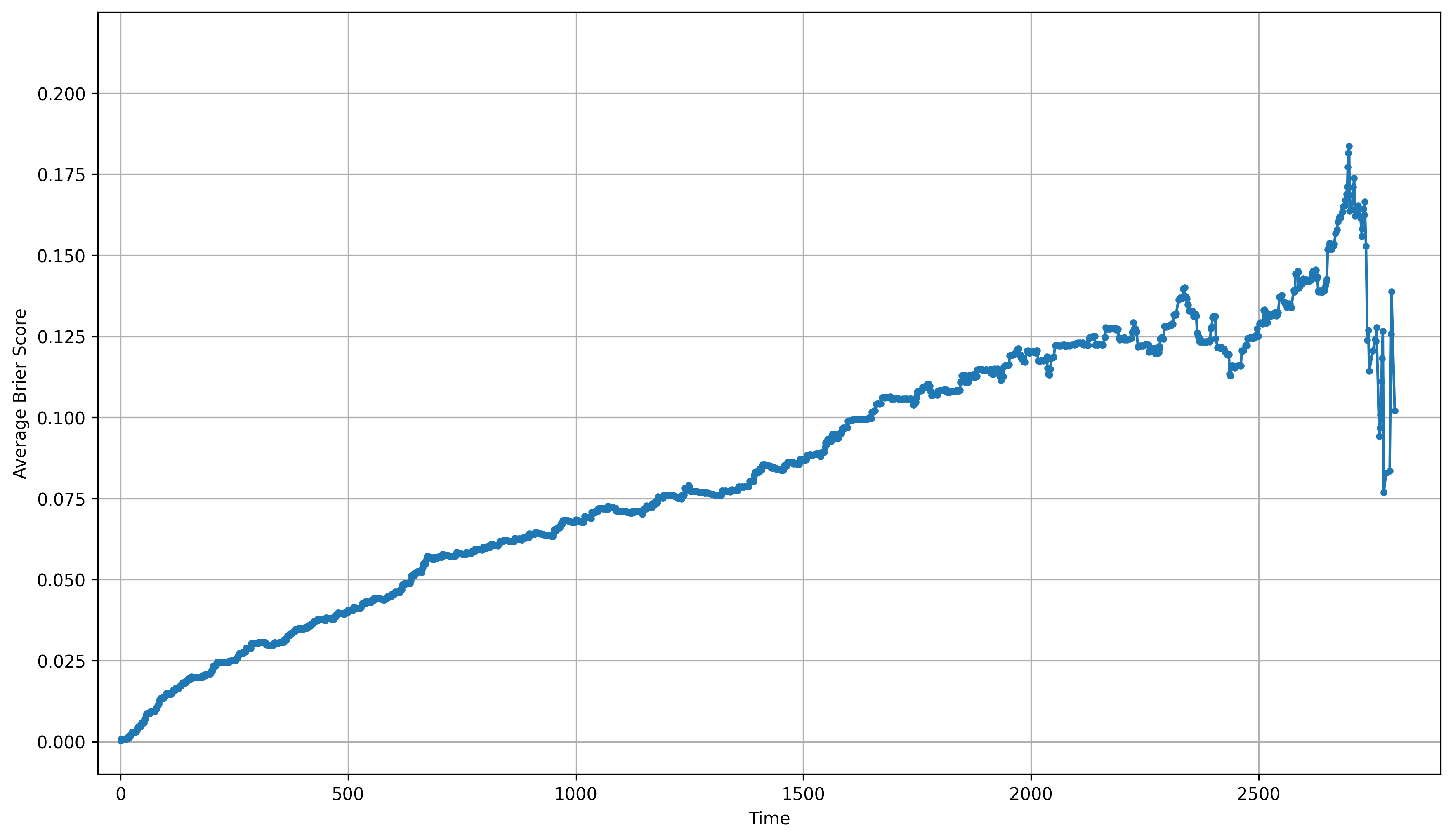}}
\end{figure}

\pagebreak

%% file: 9-appendix-H-dynamic-auc.tex
\section{Dynamic AUC}\label{apd:time_dep_auc}

\begin{figure}[htbp]
    \floatconts
    {fig:baseline_time_dep_auc} % Label goes here
    {\caption{Time-dependent AUC plot for baseline Cox model.}}
    {\includegraphics[width=\linewidth]{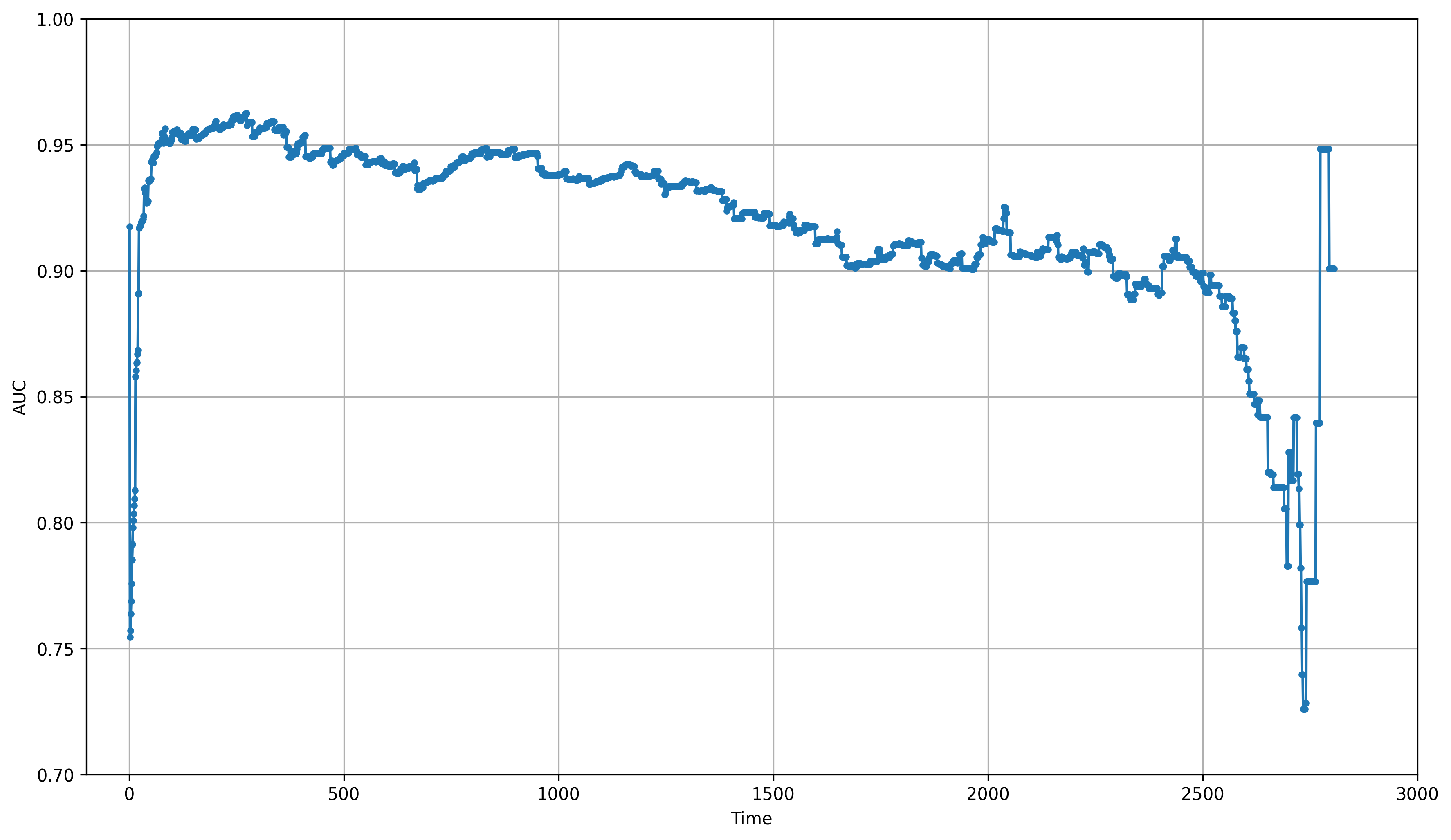}}
\end{figure}

\begin{figure}[htbp]
    \floatconts
    {fig:xgboost_time_dep_auc} % Label goes here
    {\caption{Time-dependent AUC plot for XGboost-augmented Cox model.}}
    {\includegraphics[width=\linewidth]{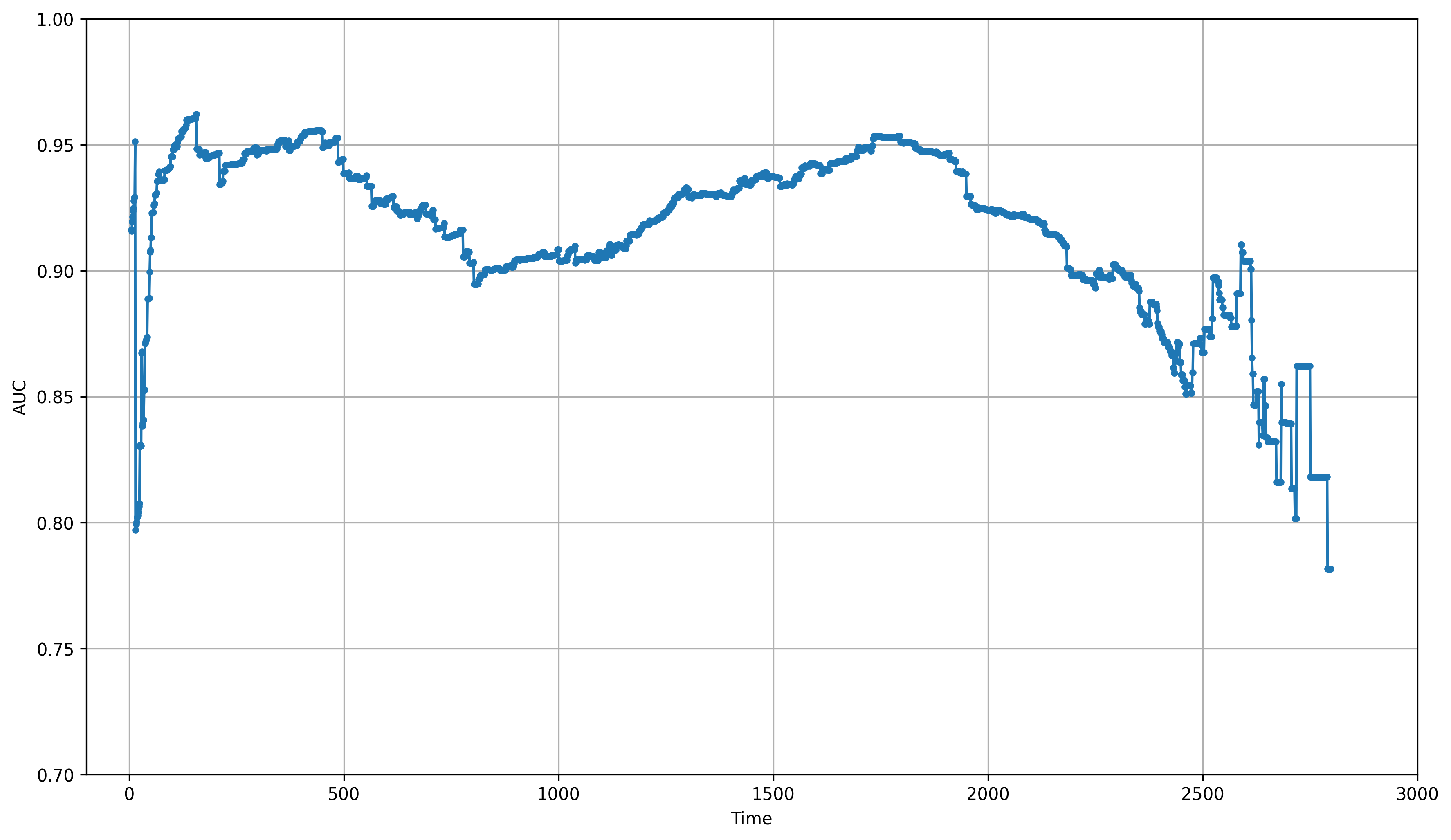}}
\end{figure}

\begin{figure}[htbp]
    \floatconts
    {fig:fcnn_time_dep_auc} % Label goes here
    {\caption{Time-dependent AUC plot for FCNN-augmented Cox model.}}
    {\includegraphics[width=\linewidth]{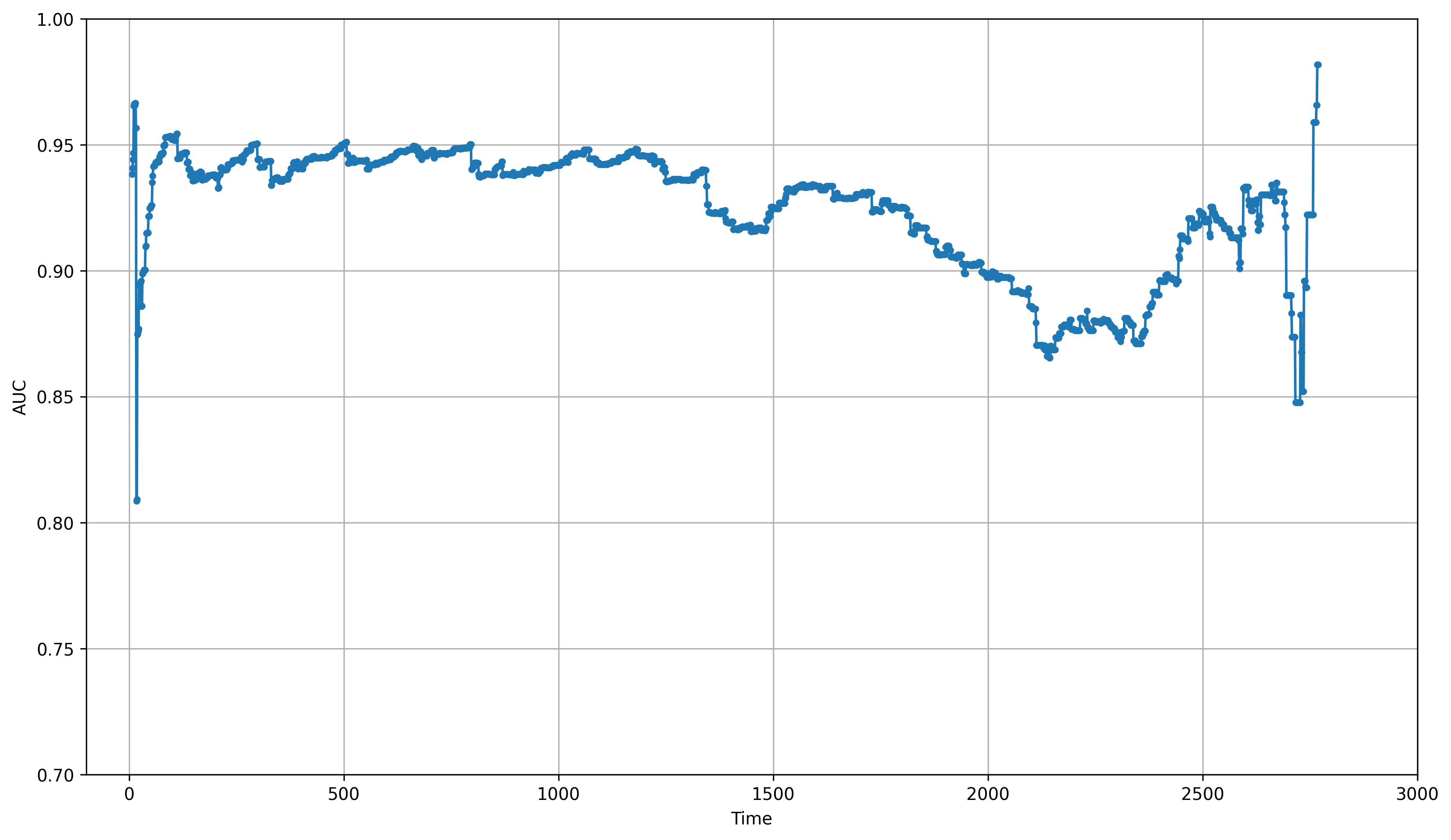}}
\end{figure}

\begin{figure}[htbp]
    \floatconts
    {fig:resnet_time_dep_auc} % Label goes here
    {\caption{Time-dependent AUC plot for ResNet-augmented Cox model.}}
    {\includegraphics[width=\linewidth]{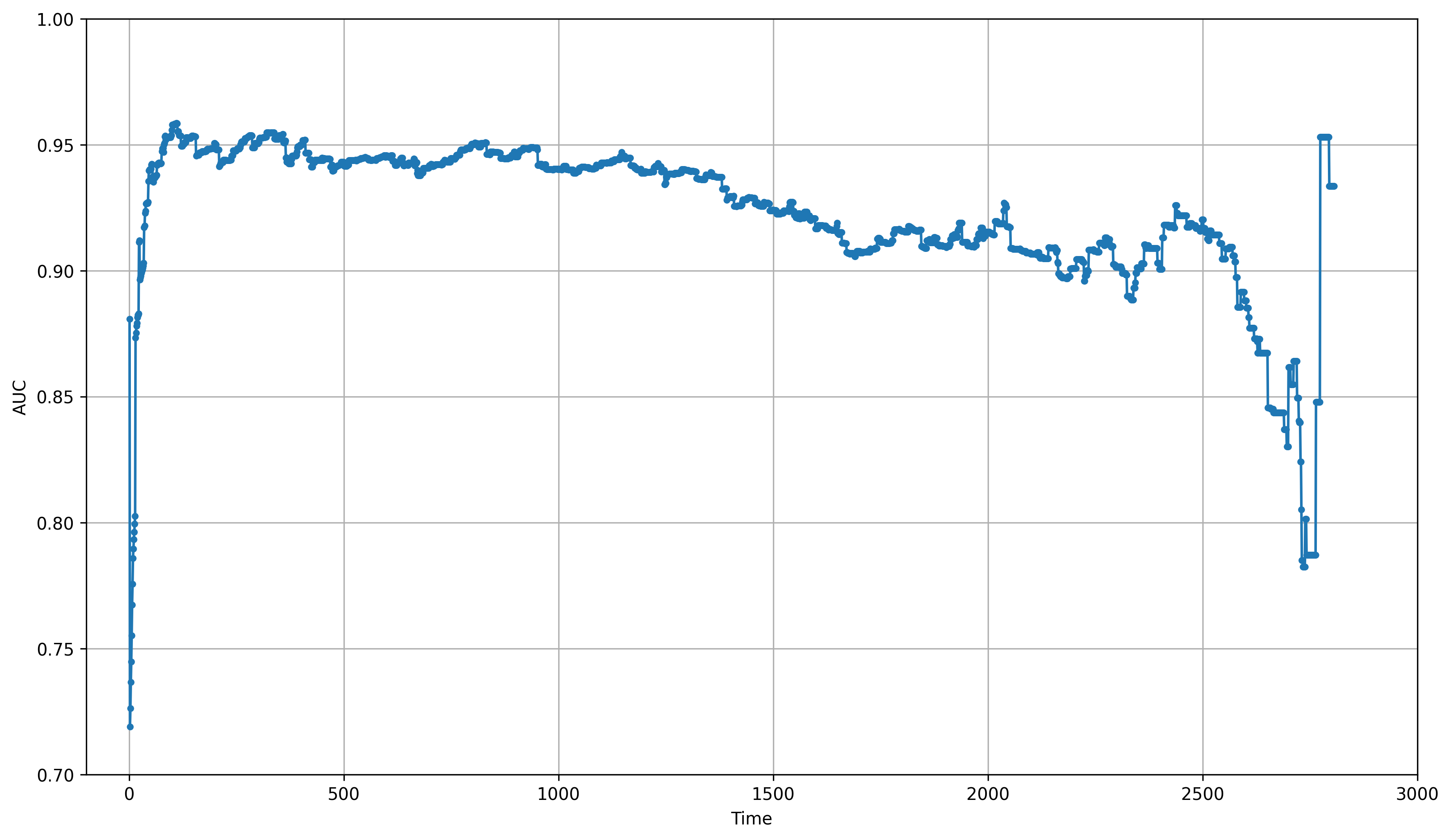}}
\end{figure}

\begin{figure}[htbp]
    \floatconts
    {fig:lr_time_dep_auc} % Label goes here
    {\caption{Time-dependent AUC plot for LR-augmented Cox model.}}
    {\includegraphics[width=\linewidth]{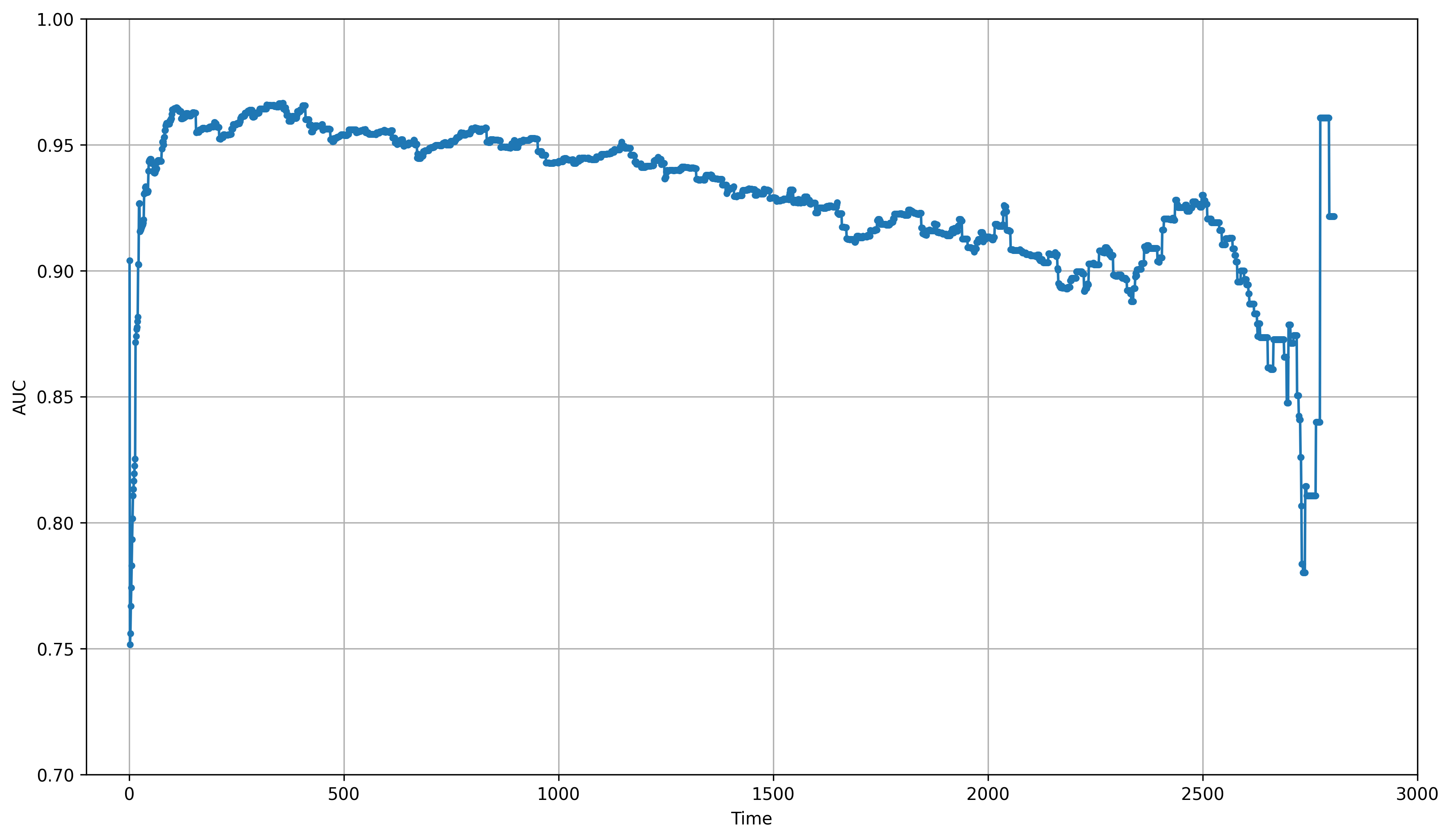}}
\end{figure}

\begin{figure}[htbp]
    \floatconts
    {fig:dt_time_dep_auc} % Label goes here
    {\caption{Time-dependent AUC plot for DT-augmented Cox model.}}
    {\includegraphics[width=\linewidth]{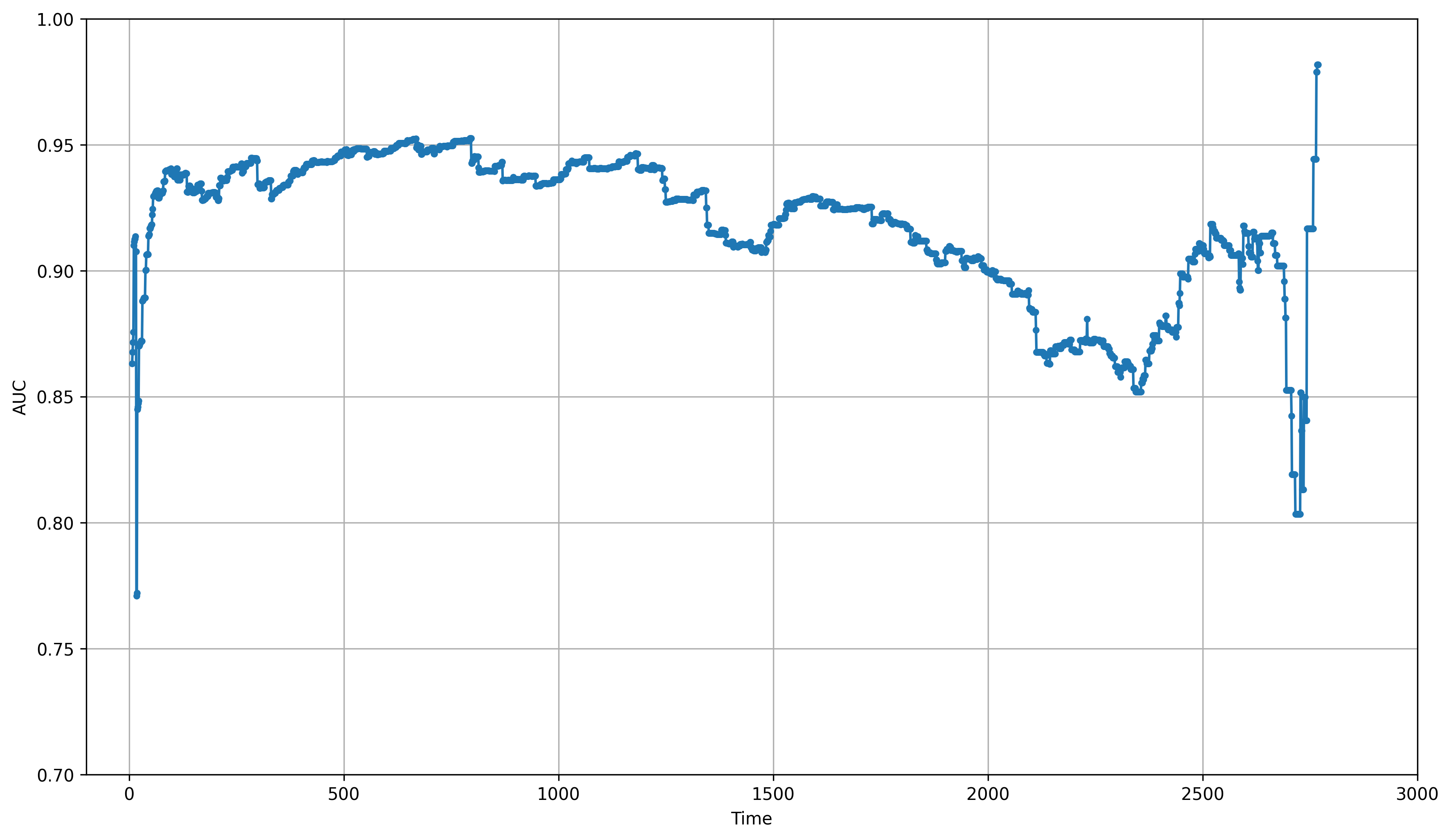}}
\end{figure}

\begin{figure}[htbp]
    \floatconts
    {fig:rf_time_dep_auc} % Label goes here
    {\caption{Time-dependent AUC plot for RF-augmented Cox model.}}
    {\includegraphics[width=\linewidth]{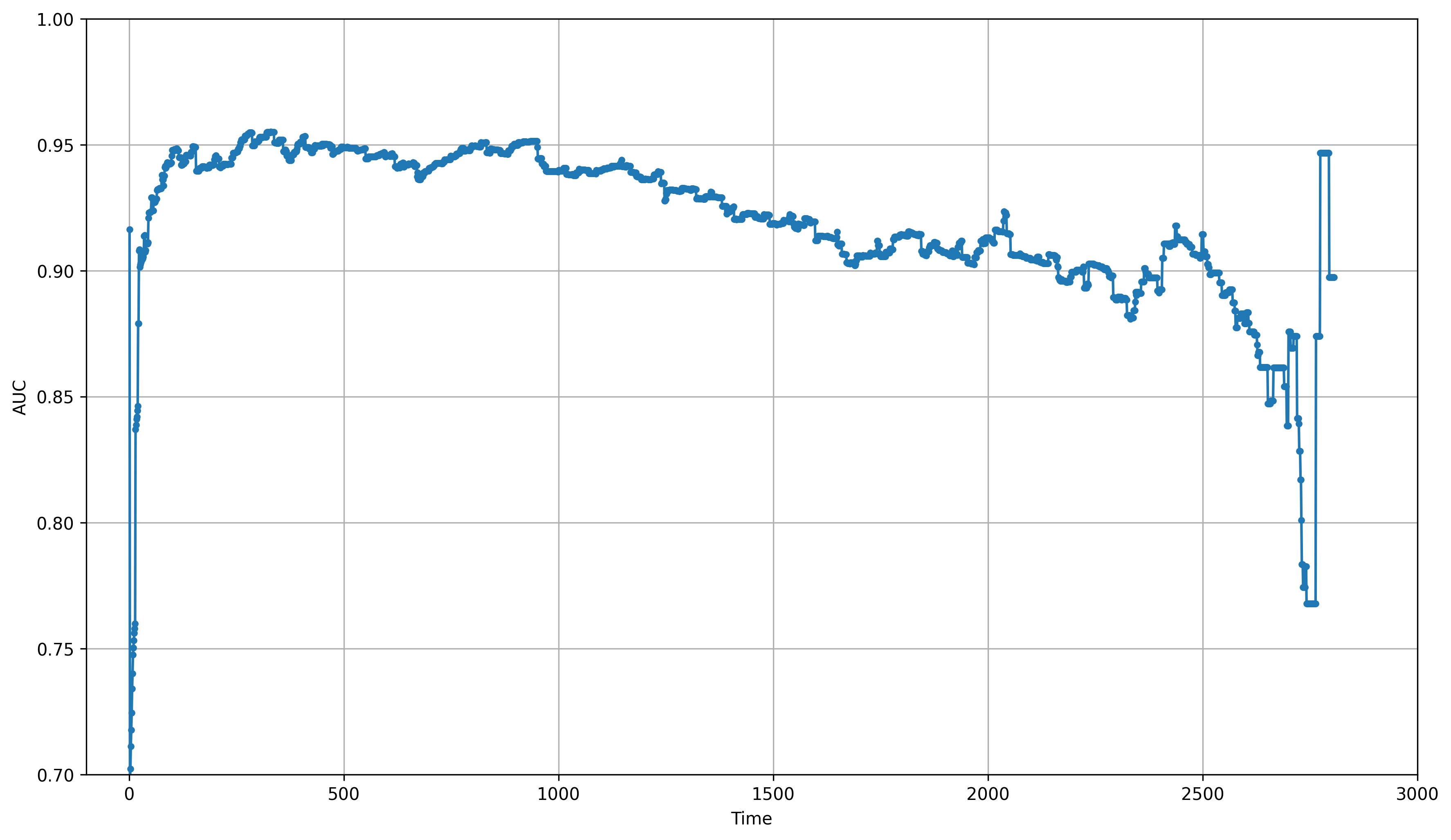}}
\end{figure}

% \section{Survival Curves}\label{apd:surv_curv}